\documentclass[11pt]{article}

\usepackage[final]{acl}

\usepackage{times}
\usepackage{latexsym}

\usepackage[T1]{fontenc}
\usepackage[utf8]{inputenc}

\usepackage{microtype}
\usepackage{inconsolata}

\usepackage{graphicx}
\usepackage{booktabs}
\usepackage{multirow}
\usepackage[table]{xcolor}
\usepackage{adjustbox}
\usepackage{tabularx}
\usepackage{array}
\usepackage{float}

\usepackage{amsmath}
\usepackage{amssymb}

\usepackage{algorithm}
\usepackage{algpseudocode}

\usepackage{enumitem}
\usepackage{pifont}

\usepackage{tikz}
\usepackage{fontawesome5}
\usepackage[nosfdefault]{comicneue}

\definecolor{bestred}{RGB}{160,85,85}
\definecolor{bestguiblue}{RGB}{31,87,150}
\definecolor{bestyellow}{RGB}{178,134,0}

\definecolor{projectlink}{RGB}{27,103,151}
\definecolor{codelink}{RGB}{36,41,47}

\newcommand{\bestvideo}[1]{%
    \textcolor{bestred}{\textbf{#1}}%
}

\newcommand{\bestgui}[1]{%
    \textcolor{bestguiblue}{\textbf{#1}}%
}

\newcommand{\bestlb}[1]{%
    \textcolor{bestyellow}{\textbf{#1}}%
}

\newcommand{\linktext}[3]{%
    \href{#2}{%
        \textcolor{#1}{%
            \comicneue
            \fontsize{10.6}{11.8}\selectfont
            #3%
        }%
    }%
}

\newcommand{\projectbutton}{%
    \linktext{projectlink}%
        {https://ryu1ion.github.io/official_BACON/}%
        {\raisebox{-0.07em}{\faGlobe}\hspace{0.38em}\textbf{Project}}%
}

\newcommand{\codebutton}{%
    \linktext{codelink}%
        {https://github.com/ryu1ion/official_BACON}%
        {\raisebox{-0.07em}{\faGithub}\hspace{0.38em}\textbf{Code}}%
}

\title{
\textit{Last But Not Least}:
\textbf{B}oundary
\textbf{A}ttention
\textbf{C}alibrati\textbf{ON}
for Multimodal KV Cache Compression
}

\author{
Tianhao Chen{$^1$}\quad
Yuheng Wu{$^1$}\quad
Kelu Yao{$^3$}\quad
Xiaogang Xu{$^4$}\\
\textbf{Xiaobin Hu}{$^{2,\ddagger}$}\quad
\textbf{Dongman Lee}{$^{1,\ddagger}$}\\[0.25em]
{$^1$} KAIST \quad
{$^2$} National University of Singapore\\
{$^3$} Zhejiang Laboratory \quad
{$^4$} The Chinese University of Hong Kong\\[0.48em]
\projectbutton
\hspace{2em}
\codebutton
}

\begin{document}

\maketitle

% =========================================================
% Corresponding authors
% =========================================================
\begingroup
\renewcommand\thefootnote{}
\footnotetext{$^\ddagger$ Corresponding authors.}
\addtocounter{footnote}{-1}
\endgroup

% =========================================================
% Abstract
% =========================================================
\begin{abstract}
Multimodal Large Language Models (MLLMs) achieve strong vision-language reasoning but incur large KV caches and high decoding latency with long visual contexts. Existing compression methods rely on observation window attention for stable token importance estimation, yet this aggregation can dilute sparse critical evidence and discard answer-relevant tokens under aggressive compression. We identify last query attention as a complementary signal for recovering such evidence, though its irrelevant signals may introduce additional noise. We propose BACON, a plug-and-play method that calibrates observation window attention with last query evidence while suppressing noise through intra-layer coherence and inter-layer persistence. Across diverse benchmarks, models, budgets, and compression methods, BACON improves multimodal KV-cache compression by 7.5\% on average under the most aggressive budget, with gains up to 30.9\%.
\end{abstract}

% =========================================================
% Main content
% =========================================================
\section{Introduction}
Multimodal Large Language Models (MLLMs) have become a central foundation for vision-language intelligence, extending the reasoning and generation capabilities of Large Language Models (LLMs) to multimodal scenarios through visual encoders and projection modules. Despite strong performance across multimodal tasks, efficient inference remains challenging as visual inputs become longer and denser. Visual tokens substantially increase the prefill length \citep{kwon2023efficient}, and the KV cache must retain their representations across layers and attention heads for subsequent decoding. KV cache compression has therefore become an important direction for reducing the memory and latency cost of MLLM inference. 
\begin{figure}[t]
    \centering
    \includegraphics[width=\columnwidth]{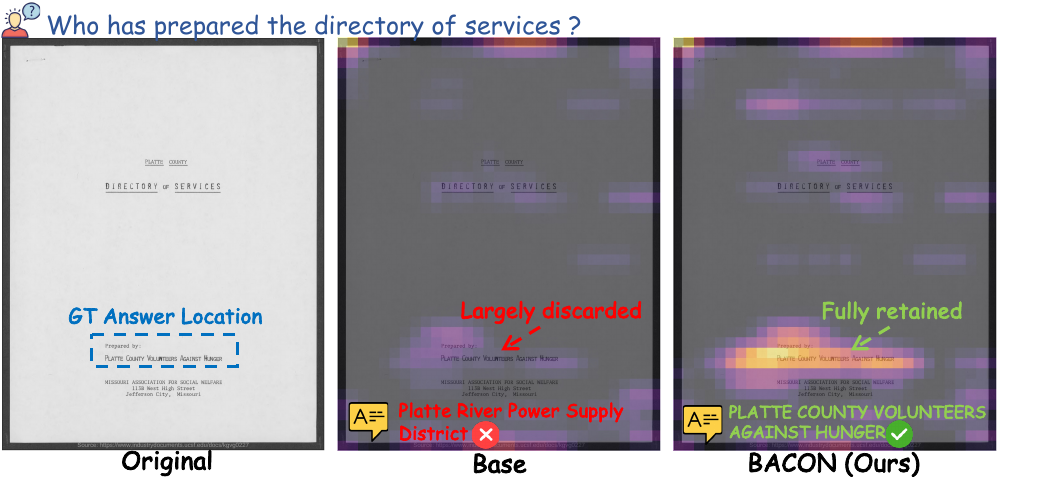}
    \vspace{-9mm}
    \caption{Visual Importance Estimation under low budget on SnapKV: Observation window vs. BACON}
    \label{fig:visualization}
    \vspace{-8.5mm}
\end{figure}
\begin{figure*}[!t]
\centering
\includegraphics[width=\textwidth]{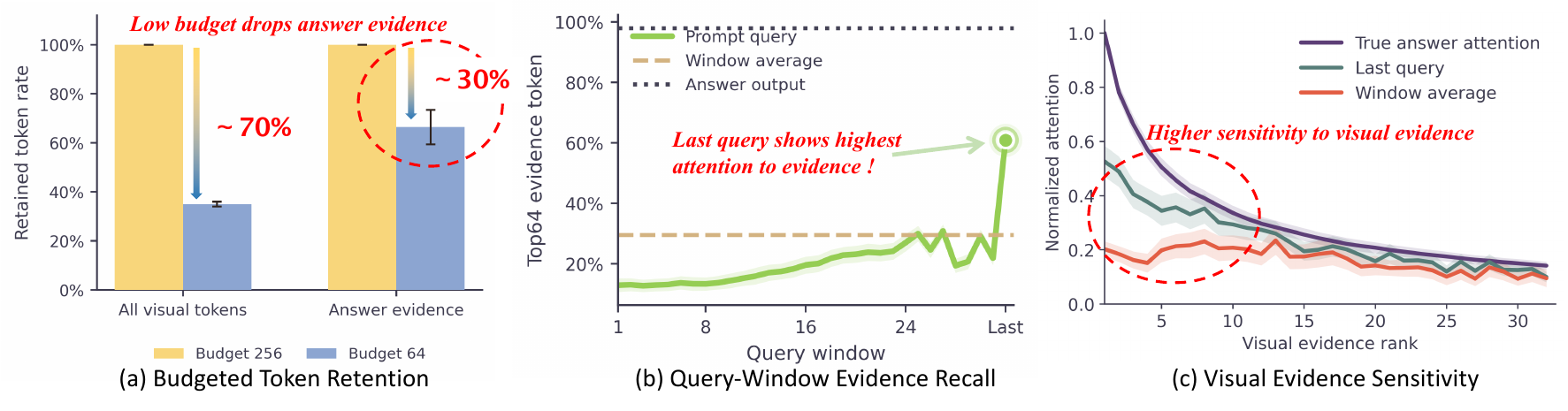}
\vspace{-9mm}
\caption{
\textbf{Observation window aggregation can dilute sparse visual evidence, while the last query recovers these evidence.}
(a) Under aggressive KV compression, SparseMM discard answer-critical visual evidence and produce incorrect predictions, suggesting observation window attention can miss sparse evidence under tight cache budgets.
(b) The last query is more sensitive than earlier prompt queries to answer-relevant tokens, showing its ability to capture boundary-emergent evidence.
(c) Compared with window-averaged attention, last-query attention better highlights visually important tokens and closely matches the answer's attention distribution over visual tokens.
}
\label{fig:motivation1}
\vspace{-5mm}
\end{figure*}

Existing KV cache compression methods typically retain cached tokens according to their estimated importance.
A common strategy is to estimate this importance from an observation window \citep{li2024snapkv}, i.e., the final segment of prompt queries, by averaging the attention each cached token receives from these queries.
While this averaging reduces query-specific noise and stabilizes importance estimation, it also introduces a fundamental limitation: sparse answer-relevant visual evidence can be obscured by query-dependent attention signals unrelated to the target answer~\citep{kang2025see}.
Consequently, \textbf{window aggregation can dilute sparse visual evidence and bias retention away from answer-critical visual tokens}, especially under aggressive compression, as illustrated in Fig.~\ref{fig:visualization} and  Fig.~\ref{fig:motivation1}(a).

\begin{figure*}[!t]
\centering
\includegraphics[width=\textwidth]{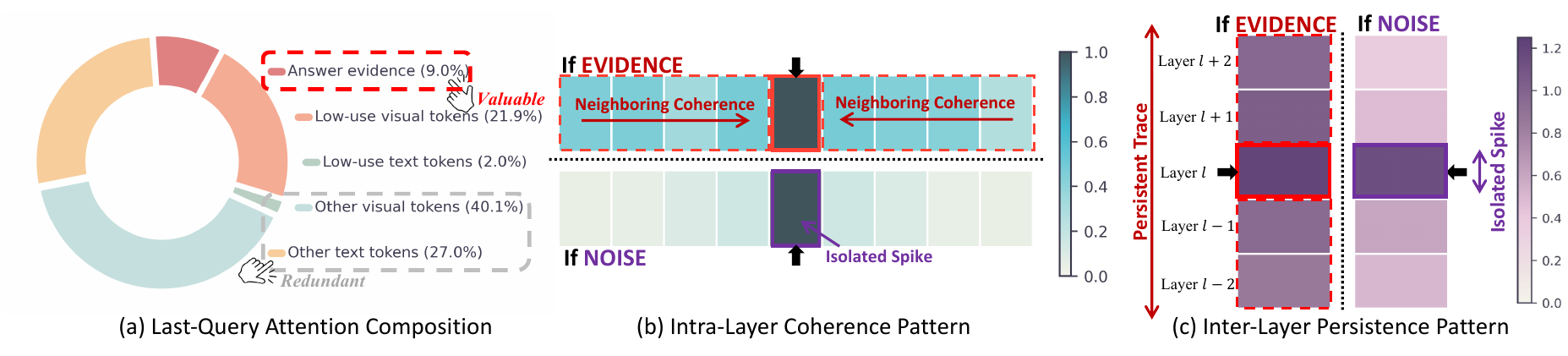}
\vspace{-9mm}
\caption{
\textbf{Why last-query attention needs calibration.}
(a) The last query can highlight answer evidence, but most of its high-attention tokens are not useful: only about 9\% correspond to truly important evidence, while around 70\% are answer-irrelevant noise.
(b) Important visual evidence usually appears as a local region, where neighboring tokens also receive high attention; in contrast, noise often appears as an isolated attention spike.
(c) Important visual evidence also remains salient across adjacent layers, while noisy spikes are less consistent.
}
\label{fig:motivation2}
\vspace{-4.5mm}
\end{figure*}

We investigate this limitation by analyzing attention dynamics in MLLM KV cache compression.
Our analysis in Fig.~\ref{fig:motivation1}(b) and~\ref{fig:motivation1}(c) shows that the last prompt query can reveal important boundary evidence weakened by observation window aggregation.
However, it is not sufficiently reliable as a standalone retention criterion: as shown in Fig.~\ref{fig:motivation2}(a), nearly 70\% of its high-attention tokens are non-evidential.
Directly incorporating last-query attention into token retention can therefore assign excessive importance to answer-irrelevant high-attention tokens, while weakening the stability provided by observation window aggregation.
This leads to a key insight: \textit{the last query is not least for token retention; it reveals visual evidence missed by observation window aggregation, but this evidence should be calibrated against the window-based signal before guiding token retention.} 

Motivated by this insight, we propose \textbf{B}oundary \textbf{A}ttention \textbf{C}alibrati\textbf{ON} (\textbf{BACON}), a plug-and-play token retention mechanism for MLLM KV cache compression. BACON keeps observation window attention as the stable basis of existing retention scores and uses last-query attention to calibrate it with boundary-emergent evidence. However, such boundary evidence must be distinguished from noisy high-attention signals before being directly incorporated into token retention. BACON therefore filters last-query signals through local coherence within each layer and persistence across adjacent layers, as discovered in Fig.~\ref{fig:motivation2}(b) and~\ref{fig:motivation2}(c). In this way, BACON recovers visual evidence diluted by window aggregation while avoiding the noise introduced by directly using last-query attention. As shown in Fig.~\ref{fig:visualization}, BACON successfully captures key visual information needed for questions compared to base counterparts. More visualization results are available in \ref{append:vis}.

In summary, our contributions are as follows. (1) We identify observation window aggregation as a key limitation why sparse but answer-critical visual evidence can be diluted and discarded during aggressive KV cache compression. (2) We propose BACON, a plug-and-play boundary attention calibration method that leverages evidence revealed by the last query to calibrate stable but diluted observation window attention, while suppressing isolated noise through intra-layer coherence and inter-layer persistence. (3) We conduct extensive experiments spanning multimodal understanding, video reasoning, GUI grounding, and long-context text tasks, with (M)LLMs of different scales and architectures, multiple KV compression methods, and diverse cache budgets, showing that BACON consistently improves existing compression methods without introducing extra inference cost.
\section{Related Work}
\label{sec:related_work}

\noindent\textbf{Multimodal Large Language Models.}
Recent Multimodal Large Language Models (MLLMs) extend LLMs with visual encoders and projection modules, aligning visual features with the language embedding space for joint vision-language reasoning. Representative models, such as LLaVA \citep{liu2024llavanext}, InternVL \citep{zhu2025internvl3}, and Qwen3-VL \citep{bai2025qwen3}, have achieved strong performance across diverse multimodal tasks. Recent advances further support high-resolution inputs \citep{guo2024llava}, dynamic image tiling \citep{li2024llava}, and multi-frame video reasoning \citep{maaz2024video}, improving fine-grained perception and long-context multimodal understanding. However, the resulting growth in visual context length substantially increases KV cache memory and decoding latency, making efficient inference critical for scalable deployment.

\noindent\textbf{Efficient Inference and KV Cache Compression in MLLMs.}
Efficient MLLM inference has been explored mainly through visual-token reduction and KV cache compression. Visual-token reduction methods prune intermediate visual tokens \citep{chen2024image}, merge redundant tokens \citep{shang2025llava}, or adapt input resolution \citep{wang2024qwen2}, reducing computation but often modifying the visual processing pipeline or effective visual context. In contrast, KV cache compression reduces memory and decoding cost after prefill while preserving the original input representation and forward pipeline \citep{kim2026kvzip,fu2025not}. In LLMs, StreamingLLM \citep{xiao2024efficient} and H2O \citep{zhang2023h2o} motivate retention through attention sinks and heavy hitters, while SnapKV \citep{li2024snapkv}, PyramidKV \citep{cai2024pyramidkv}, and AdaKV \citep{feng2026ada} rely on observation window importance estimation and non-uniform layer/head budgets. Recent MLLM-oriented methods further adapt cache compression to multimodal inference \citep{wan2024look}: SparseMM \citep{wang2025sparsemm} allocates modality-sensitive budgets, InfiniPotV \citep{kim2026infinipot} uses value-norm-based visual KV selection, and MixKV \citep{liu2025mixing} balances attention-based importance with semantic diversity.
Despite these advances, most attention-based methods still use observation window attention as the primary retention signal. BACON is orthogonal to these methods: it calibrates window-based retention with last-query boundary evidence while preserving the original budget allocation and decoding pipeline.
\section{Motivation Study}
\label{sec:motivation}

In this section, we analyze the attention patterns and paired failure cases of Qwen2-VL under SparseMM compression on DocVQA and present our key findings below:

\noindent\textbf{\ding{202} Low-budget compression discards visual evidence.}
To understand failures under tight cache budgets, we compare paired cases in which SparseMM succeeds with a larger budget but fails with a smaller budget. As shown in Fig.~\ref{fig:motivation1}(a), the low-budget setting removes nearly 70\% of the visual tokens retained by the high-budget setting and discards approximately 30\% of the answer-aligned evidence. And many of the removed evidence tokens correspond to low-attention visual regions that are still essential for grounding the answer. This finding indicates that aggressive compression can eliminate not only redundant visual context but also weak yet semantically decisive evidence. As visual tokens generally receive lower attention than text tokens~\citep{chen2024image}, they are especially vulnerable under low-budget selection, which motivates a retention signal beyond raw observation window attention.

\noindent\textbf{\ding{203} Last query captures boundary evidence.}
We next examine where the missing evidence appears in the attention dynamics by comparing prompt queries at different positions with observation window attention.
Fig.~\ref{fig:motivation1}(b) shows that earlier prompt queries cover only limited answer-aligned evidence, whereas the last query captures a much larger portion of tokens that are important for the answer. Fig.~\ref{fig:motivation1}(c) further shows that this advantage is especially clear for visual tokens: last-query attention assigns higher saliency to visually important tokens and more closely matches the answer's attention distribution over visual tokens than observation window attention.
These results suggest that while observation window aggregation can dilute sparse visual evidence, the last query remains sensitive to boundary-emergent visual signals that are closely tied to the forthcoming answer.

\noindent\textbf{\ding{204} Intra-layer coherence and inter-layer persistence calibrate noisy boundary evidence.}
Although last-query attention can discover boundary evidence weakened by observation window aggregation, Fig.~\ref{fig:motivation2}(a) shows that its high-attention tokens are highly mixed: only 9\% correspond to true answer evidence, whereas nearly 70\% are answer-irrelevant noise.
Directly using raw last-query attention for retention would therefore introduce irrelevant signals that interfere with important boundary evidence.
As shown in Figs.~\ref{fig:motivation2}(b) and \ref{fig:motivation2}(c), true evidence exhibits stronger intra-layer coherence and inter-layer persistence: high attention extends from evidence tokens to neighboring tokens within the same layer and remains concentrated on the same token positions across adjacent layers.
In contrast, answer-irrelevant high-attention tokens are more isolated within a layer and less stable across layers.
These observations suggest that important boundary evidence should be salient to the last query while at the same time exhibiting intra-layer coherence and inter-layer persistence.
\newcommand{\Std}{\mathrm{Std}}
\section{Methodology}
\label{sec:methodology}
\vspace{-5pt}

\subsection{Preliminaries: Attention-Based KV Cache Compression}

Given a multimodal prompt $\mathbf{X}=\{\mathbf{x}_i\}_{i=1}^{N}$, an MLLM computes the KV cache during prefill and reuses it during autoregressive decoding. For layer $l$ and attention head $h$, we denote the cached keys and values by $\mathbf{K}^{l}_{h},\mathbf{V}^{l}_{h}\in\mathbb{R}^{N\times d}$, where $N$ is the sequence length and $d$ is the head dimension. KV cache compression retains a compact subset of KV pairs under a budget $K_{l,h}$ by ranking candidate tokens for each layer-head pair.

\begin{figure}[t]
    \centering
    \includegraphics[width=\columnwidth]{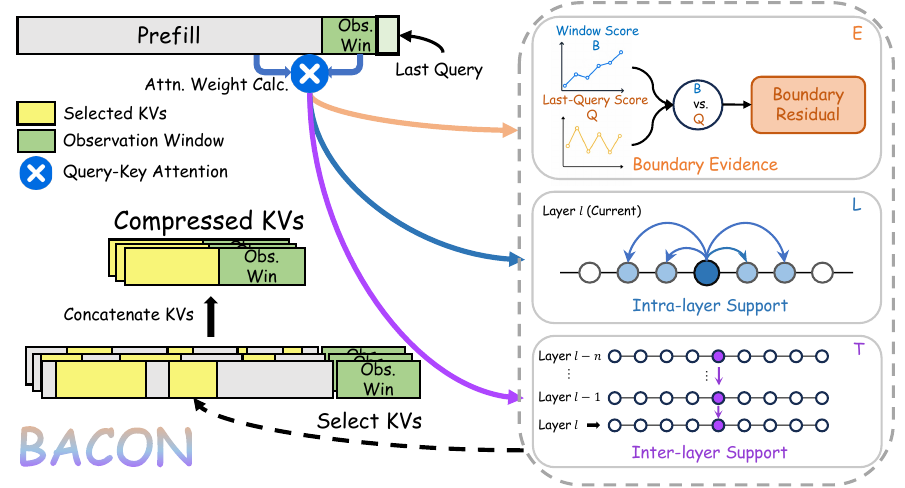}
    \vspace{-7mm}
    \caption{Overview of BACON. BACON extracts boundary evidence from observation window and last-query attention, then calibrates it with intra-layer coherence and inter-layer persistence to produce an evidence-aware score for head-wise KV cache compression.}
    \label{fig:bacon}
    \vspace{-5mm}
\end{figure}

Most attention-based methods estimate token importance by averaging attention over an observation window $\mathcal{W}$ near the end of the prompt. Let $\mathbf{A}^{l,h}\in\mathbb{R}^{N\times N}$ denote the causal attention matrix, where $A^{l,h}_{q,i}$ is the attention weight from query position $q$ to key position $i$. The observation window score $B^{l,h}_i$ is
\begin{equation}
B^{l,h}_i
=
\frac{1}{|\mathcal{W}|}
\sum_{q\in\mathcal{W}}
A^{l,h}_{q,i}.
\label{eq:window_score}
\end{equation}
Although $B^{l,h}_i$ provides a stable relevance estimate, Sec.~\ref{sec:motivation} shows that observation window aggregation can dilute sparse but answer-critical visual evidence. BACON therefore uses $B^{l,h}_i$ as the backbone score and calibrates it with boundary-emergent evidence exposed by the last query. Fig.~\ref{fig:bacon} gives an overview of BACON.

\subsection{Boundary-Emergent Evidence Modeling}
\label{subsec:boundary_modeling}

BACON is motivated by the observation in Sec.~\ref{sec:motivation} that the last prompt query can expose boundary-emergent evidence weakened by observation window aggregation, while raw last-query attention may also contain answer-irrelevant spikes. Let the last query position in the prompt be $D$. We define the last-query attention score as
\begin{equation}
Q^{l,h}_i
=
A^{l,h}_{D,i}.
\label{eq:last_query_score}
\end{equation}
We decompose last-query saliency into a stable window signal, a boundary-emergent evidence component, and a noise component:
\begin{equation}
Q^{l,h}_i
=
B^{l,h}_i
+
\Delta^{l,h}_i
+
\xi^{l,h}_i,
\label{eq:score_decomposition}
\end{equation}
where $\Delta^{l,h}_i$ denotes evidence revealed by the last prompt query, and $\xi^{l,h}_i$ denotes non-evidential noise. Since $\Delta^{l,h}_i$ is not directly observable, BACON estimates it as the boundary evidence estimate $E^{l,h}_i$ using the positive gap between last-query attention and observation window attention:
\begin{equation}
E^{l,h}_i
=
\left[
Q^{l,h}_i
-
B^{l,h}_i
\right]_+,
\label{eq:boundary_residual}
\end{equation}
where $[x]_+=\max(x,0)$. 
The positive residual $E^{l,h}_i$ measures how much more token $i$ is attended by the last query than by the observation window, thereby identifying evidence that may be diluted in the window-based score while preserving $B^{l,h}_i$ as the retention basis.

\subsection{Structural Calibration of Boundary Evidence}
\label{subsec:structural_calibration}

As shown in Sec.~\ref{sec:motivation}, useful boundary evidence tends to exhibit intra-layer coherence and inter-layer persistence. BACON therefore refines the boundary residual with these two structural cues. For each layer-head pair, we write the residual scores as $\mathbf{e}^{l,h}\in\mathbb{R}^{N}$, whose $i$-th entry is $E^{l,h}_i$.

\noindent \textbf{Intra-layer coherence.}
BACON measures intra-layer coherence by aggregating boundary residuals over neighboring token positions. For token $i$, we define a radius-$r$ neighborhood, with $r=5$ by default:
\begin{equation}
\mathcal{N}_r(i)
=
\{u \mid |u-i|\leq r,\ 1\leq u\leq N\}.
\end{equation}
The row-normalized neighborhood operator $\mathbf{P}_r\in\mathbb{R}^{N\times N}$ is
\begin{equation}
(\mathbf{P}_r)_{i,u}
=
\frac{
\mathbf{1}\{u\in\mathcal{N}_r(i)\}
}{
|\mathcal{N}_r(i)|
}.
\label{eq:neighbor_matrix}
\end{equation}
The intra-layer coherence score is
\begin{equation}
\mathbf{l}^{l,h}
=
\mathbf{P}_r\mathbf{e}^{l,h},
\qquad
L^{l,h}_i
=
(\mathbf{l}^{l,h})_i.
\label{eq:intra_layer_coherence}
\end{equation}

\noindent \textbf{Inter-layer persistence.}
BACON measures inter-layer persistence by tracing boundary residuals at the same token position across preceding layers. For layer $l$, the causal layer set is
\begin{equation}
\mathcal{P}_m(l)
=
\{j \mid \max(1,l-m)\leq j < l\},
\end{equation}
where $m$ is the persistence depth, with $m=4$ by default. The inter-layer persistence score is
\begin{equation}
T^{l,h}_i
=
\frac{1}{|\mathcal{P}_m(l)|}
\sum_{j\in\mathcal{P}_m(l)}
E^{j,h}_i.
\label{eq:inter_layer_persistence}
\end{equation}
If $\mathcal{P}_m(l)=\emptyset$, we set $T^{l,h}_i=0$; using only preceding layers keeps the formulation causal and compatible with layer-by-layer compression.

% Required packages:
% \usepackage{booktabs}
% \usepackage[table]{xcolor}
% \usepackage{adjustbox}

\definecolor{baconrow}{RGB}{239,248,244}
\definecolor{mixrow}{RGB}{248,248,248}
\definecolor{modelrow}{RGB}{232,232,232}
\definecolor{fullkv}{RGB}{112,112,112}
\definecolor{bestgreen}{RGB}{22,120,72}

\definecolor{gainc}{RGB}{22,120,72}
\definecolor{dropc}{RGB}{120,120,120}
\definecolor{samec}{RGB}{120,120,120}

% Delta marks follow the compact style: value + colored mini arrow/number.
% The table uses a 1.5pt base font, so ordinary \footnotesize or math arrows are visually too large.
% We shrink only the arrow glyph and keep the delta number slightly smaller than the cell value.
\newcommand{\gain}[2]{#1 {\textcolor{gainc}{\raisebox{0.05ex}{\scalebox{0.38}{$\scriptstyle\uparrow$}}{\fontsize{1.85pt}{1.85pt}\selectfont #2}}}}
\newcommand{\drop}[2]{#1 {\textcolor{dropc}{\raisebox{0.05ex}{\scalebox{0.38}{$\scriptstyle\downarrow$}}{\fontsize{1.85pt}{1.85pt}\selectfont #2}}}}
\newcommand{\same}[2]{#1 {\textcolor{samec}{\raisebox{0.05ex}{\scalebox{0.34}{$\scriptstyle\rightarrow$}}{\fontsize{1.85pt}{1.85pt}\selectfont #2}}}}

\newcommand{\best}[1]{\textcolor{bestgreen}{\textbf{#1}}}

\begin{table*}[t]
\centering
\caption{\textbf{Main results on \best{image understanding benchmarks.}} SparseMM is excluded in InternVL3-8B and Qwen3-VL-30B-A3B settings due to the unreleased profiling file. Arrows show deltas over Base, and ``Full KV'' denotes caching all KV pairs as the upper bound.}
\label{tab:main_results_compact}
\vspace{-3mm}
{\fontsize{3pt}{3.2pt}\selectfont
\setlength{\tabcolsep}{1.3pt}
\renewcommand{\arraystretch}{0.72}
\setlength{\aboverulesep}{0.10pt}
\setlength{\belowrulesep}{0.10pt}
\setlength{\cmidrulesep}{0.05pt}
\setlength{\heavyrulewidth}{0.45pt}
\setlength{\lightrulewidth}{0.25pt}
\setlength{\cmidrulewidth}{0.25pt}
\begin{adjustbox}{width=\textwidth}
\begin{tabular}{@{}ll*{15}{c}@{}}
\toprule
\textbf{Method} & \textbf{Variant}
 & \multicolumn{3}{c}{\textbf{DocVQA (\%)}} 
 & \multicolumn{3}{c}{\textbf{TextVQA (\%)}} 
 & \multicolumn{3}{c}{\textbf{ChartQA (\%)}} 
 & \multicolumn{3}{c}{\textbf{MMMU (\%)}} 
 & \multicolumn{3}{c}{\textbf{TextCaps}} \\
\cmidrule(lr){3-5}
\cmidrule(lr){6-8}
\cmidrule(lr){9-11}
\cmidrule(lr){12-14}
\cmidrule(lr){15-17}
& & \textbf{256} & \textbf{128} & \textbf{64} 
& \textbf{256} & \textbf{128} & \textbf{64} 
& \textbf{256} & \textbf{128} & \textbf{64} 
& \textbf{256} & \textbf{128} & \textbf{64} 
& \textbf{256} & \textbf{128} & \textbf{64} \\
\midrule

\rowcolor{modelrow}\multicolumn{17}{c}{\textbf{Qwen2-VL-7B-Instruct}} \\
\textcolor{fullkv}{\textbf{Full KV}} & \textcolor{fullkv}{Upper} 
& \multicolumn{3}{c}{\textcolor{fullkv}{93.7}} 
& \multicolumn{3}{c}{\textcolor{fullkv}{82.8}} 
& \multicolumn{3}{c}{\textcolor{fullkv}{71.3}} 
& \multicolumn{3}{c}{\textcolor{fullkv}{49.9}} 
& \multicolumn{3}{c}{\textcolor{fullkv}{1.473}} \\
\midrule

\textbf{SnapKV} & Base 
& 88.6 & 82.1 & 70.1 
& 80.6 & 77.0 & 70.3 
& 70.0 & 69.6 & 66.2 
& 49.9 & 49.8 & 49.6 
& 1.361 & 1.141 & 0.787 \\
\rowcolor{mixrow} 
& +\texttt{MixKV} 
& 91.8 & 83.9 & 70.9 
& 82.6 & 81.0 & 73.5 
& 70.0 & 70.2 & 67.2 
& 49.9 & 49.9 & 49.7 
& 1.470 & 1.332 & 0.919 \\
\rowcolor{baconrow} 
& +\texttt{BACON} 
& \best{\gain{93.1}{4.5}} & \best{\gain{91.5}{9.4}} & \best{\gain{85.5}{15.4}} 
& \best{\gain{83.1}{2.5}} & \best{\gain{82.6}{5.6}} & \best{\gain{78.2}{7.9}} 
& \best{\gain{70.2}{0.2}} & \best{\gain{70.2}{0.6}} & \best{\gain{69.6}{3.4}} 
& \best{\same{49.9}{0.0}} & \best{\gain{49.9}{0.1}} & \best{\gain{50.0}{0.4}} 
& \best{\gain{1.488}{0.127}} & \best{\gain{1.426}{0.285}} & \best{\gain{1.178}{0.391}} \\
\addlinespace[0.05pt]

\textbf{PyramidKV} & Base 
& 83.4 & 75.6 & 60.5 
& 77.7 & 74.9 & 66.8 
& 71.1 & 68.9 & 65.2 
& 49.9 & 49.8 & 49.6 
& 1.147 & 0.993 & 0.600 \\
\rowcolor{mixrow} 
& +\texttt{MixKV} 
& 85.0 & 77.5 & 61.3 
& 80.9 & 77.2 & 69.4 
& 71.0 & \best{71.1} & 66.4 
& 49.9 & 49.8 & 49.6 
& 1.383 & 1.145 & 0.662 \\
\rowcolor{baconrow} 
& +\texttt{BACON} 
& \best{\gain{92.3}{8.9}} & \best{\gain{89.3}{13.7}} & \best{\gain{79.2}{18.7}} 
& \best{\gain{82.5}{4.8}} & \best{\gain{80.0}{5.1}} & \best{\gain{75.1}{8.3}} 
& \best{\gain{71.2}{0.1}} & \gain{70.6}{1.7} & \best{\gain{69.2}{4.0}} 
& \best{\same{49.9}{0.0}} & \best{\gain{49.9}{0.1}} & \best{\gain{49.8}{0.2}} 
& \best{\gain{1.461}{0.314}} & \best{\gain{1.344}{0.351}} & \best{\gain{1.073}{0.473}} \\
\addlinespace[0.05pt]
\textbf{AdaKV} & Base 
& 88.4 & 81.3 & 69.4 
& 80.5 & 75.9 & 70.8 
& 69.8 & 69.6 & 66.6 
& 49.9 & 49.7 & 49.6 
& 1.300 & 1.099 & 0.771 \\
\rowcolor{mixrow} 
& +\texttt{MixKV} 
& 91.4 & 82.7 & 70.7 
& 82.5 & 79.1 & 72.6 
& 70.2 & 70.2 & 67.8 
& 49.9 & \best{49.9} & 49.6 
& 1.454 & 1.271 & 0.874 \\
\rowcolor{baconrow} 
& +\texttt{BACON} 
& \best{\gain{93.0}{4.6}} & \best{\gain{91.1}{9.8}} & \best{\gain{86.2}{16.8}} 
& \best{\gain{82.8}{2.3}} & \best{\gain{81.2}{5.3}} & \best{\gain{78.5}{7.7}} 
& \best{\gain{70.2}{0.4}} & \best{\gain{70.2}{0.6}} & \best{\gain{69.6}{3.0}} 
& \best{\same{49.9}{0.0}} & \gain{49.8}{0.1} & \best{\gain{49.8}{0.2}} 
& \best{\gain{1.473}{0.173}} & \best{\gain{1.371}{0.272}} & \best{\gain{1.144}{0.373}} \\
\addlinespace[0.05pt]

\textbf{SparseMM} & Base 
& 93.1 & 91.4 & 87.3 
& 82.6 & 82.1 & 76.9 
& 70.2 & 70.0 & 69.6 
& 49.8 & 49.8 & 49.6 
& 1.481 & 1.427 & 1.044 \\
\rowcolor{mixrow} 
& +\texttt{MixKV} 
& \best{93.9} & 92.9 & 88.6 
& 82.5 & \best{82.5} & 80.9 
& 69.6 & 69.8 & \best{70.8} 
& 49.8 & 49.8 & 49.7 
& 1.480 & 1.456 & 1.303 \\
\rowcolor{baconrow} 
& +\texttt{BACON} 
& \gain{93.8}{0.7} & \best{\gain{93.2}{1.8}} & \best{\gain{92.0}{4.7}} 
& \best{\same{82.6}{0.0}} & \gain{82.4}{0.3} & \best{\gain{81.6}{4.7}} 
& \best{\gain{70.6}{0.4}} & \best{\gain{70.4}{0.4}} & \gain{70.2}{0.6} 
& \best{\gain{49.9}{0.1}} & \best{\same{49.8}{0.0}} & \best{\gain{49.8}{0.2}} 
& \best{\gain{1.506}{0.025}} & \best{\gain{1.511}{0.084}} & \best{\gain{1.431}{0.387}} \\
\addlinespace[0.05pt]

\midrule
\rowcolor{modelrow}\multicolumn{17}{c}{\textbf{LLaVA-NeXT-Mistral-7B}} \\
\textcolor{fullkv}{\textbf{Full KV}} & \textcolor{fullkv}{Upper} 
& \multicolumn{3}{c}{\textcolor{fullkv}{62.7}} 
& \multicolumn{3}{c}{\textcolor{fullkv}{68.4}} 
& \multicolumn{3}{c}{\textcolor{fullkv}{51.8}} 
& \multicolumn{3}{c}{\textcolor{fullkv}{34.7}} 
& \multicolumn{3}{c}{\textcolor{fullkv}{0.704}} \\
\midrule

\textbf{SnapKV} & Base 
& 58.1 & 55.2 & 46.2 
& 66.0 & 63.0 & 58.9 
& 49.8 & 48.4 & 47.2 
& 34.7 & 34.9 & 34.8 
& 0.651 & 0.560 & 0.442 \\
\rowcolor{mixrow} 
& +\texttt{MixKV} 
& \best{60.4} & 57.5 & 48.2 
& \best{67.5} & 66.1 & 61.3 
& 49.8 & 48.6 & 46.8 
& 34.6 & 34.8 & 34.8 
& \best{0.710} & 0.656 & 0.510 \\
\rowcolor{baconrow} 
& +\texttt{BACON} 
& \gain{59.7}{1.6} & \best{\gain{57.8}{2.6}} & \best{\gain{53.9}{7.7}} 
& \gain{67.1}{1.1} & \best{\gain{66.2}{3.2}} & \best{\gain{62.8}{3.9}} 
& \best{\gain{50.0}{0.2}} & \best{\gain{49.6}{1.2}} & \best{\gain{49.2}{2.0}} 
& \best{\gain{34.8}{0.1}} & \best{\same{34.9}{0.0}} & \best{\same{34.8}{0.0}} 
& \gain{0.686}{0.035} & \best{\gain{0.676}{0.116}} & \best{\gain{0.512}{0.070}} \\
\addlinespace[0.05pt]

\textbf{PyramidKV} & Base 
& 57.5 & 54.3 & 43.8 
& 65.2 & 63.1 & 55.7 
& 43.6 & 42.7 & 38.6 
& 34.8 & 35.0 & 34.7 
& 0.652 & 0.581 & 0.436 \\
\rowcolor{mixrow} 
& +\texttt{MixKV} 
& 60.3 & 56.7 & 45.6 
& 67.2 & 65.8 & 57.8 
& 44.3 & 42.6 & 39.1 
& 34.7 & 34.9 & 34.7 
& 0.685 & 0.644 & 0.505 \\
\rowcolor{baconrow} 
& +\texttt{BACON} 
& \best{\gain{60.6}{3.1}} & \best{\gain{59.0}{4.7}} & \best{\gain{52.4}{8.6}} 
& \best{\gain{67.4}{2.2}} & \best{\gain{66.0}{2.9}} & \best{\gain{59.8}{4.1}} 
& \best{\gain{45.0}{1.4}} & \best{\gain{43.9}{1.2}} & \best{\gain{39.9}{1.3}} 
& \best{\gain{35.0}{0.2}} & \best{\same{35.0}{0.0}} & \best{\gain{34.8}{0.1}} 
& \best{\gain{0.685}{0.033}} & \best{\gain{0.650}{0.069}} & \best{\gain{0.505}{0.069}} \\
\addlinespace[0.05pt]

\textbf{AdaKV} & Base 
& 58.3 & 56.1 & 47.6 
& 65.4 & 62.7 & 57.8 
& 49.6 & 48.8 & 47.4 
& 34.7 & 34.9 & 34.8 
& 0.645 & 0.568 & 0.441 \\
\rowcolor{mixrow} 
& +\texttt{MixKV} 
& 59.3 & 57.5 & 49.5 
& \best{67.2} & 64.4 & 59.6 
& 49.6 & 49.2 & 46.8 
& 34.7 & 34.9 & 34.8 
& \best{0.701} & 0.660 & 0.506 \\
\rowcolor{baconrow} 
& +\texttt{BACON} 
& \best{\gain{59.3}{1.0}} & \best{\gain{58.4}{2.3}} & \best{\gain{55.2}{7.6}} 
& \gain{67.1}{1.7} & \best{\gain{65.5}{2.8}} & \best{\gain{61.5}{3.7}} 
& \best{\gain{50.0}{0.4}} & \best{\gain{50.2}{1.4}} & \best{\gain{49.0}{1.6}} 
& \best{\gain{34.8}{0.1}} & \best{\same{34.9}{0.0}} & \best{\same{34.8}{0.0}} 
& \gain{0.694}{0.049} & \best{\gain{0.671}{0.103}} & \best{\gain{0.510}{0.069}} \\
\addlinespace[0.05pt]

\textbf{SparseMM} & Base 
& 58.5 & 58.9 & 57.5 
& 67.2 & 67.4 & 65.2 
& 50.6 & 49.8 & 49.4 
& 34.7 & 34.7 & 34.8 
& 0.670 & 0.600 & 0.489 \\
\rowcolor{mixrow} 
& +\texttt{MixKV} 
& 58.9 & 59.0 & \best{58.7} 
& 67.4 & 67.6 & 67.1 
& \best{51.0} & 50.4 & \best{50.0} 
& 34.8 & 34.7 & 34.7 
& 0.685 & 0.620 & 0.569 \\
\rowcolor{baconrow} 
& +\texttt{BACON} 
& \best{\gain{59.7}{1.2}} & \best{\gain{59.3}{0.4}} & \gain{58.3}{0.8} 
& \best{\gain{68.0}{0.8}} & \best{\gain{67.8}{0.4}} & \best{\gain{67.3}{2.1}} 
& \same{50.6}{0.0} & \best{\gain{50.4}{0.6}} & \gain{49.8}{0.4} 
& \best{\gain{34.8}{0.1}} & \best{\gain{34.8}{0.1}} & \best{\gain{34.9}{0.1}} 
& \best{\gain{0.686}{0.016}} & \best{\gain{0.625}{0.025}} & \best{\gain{0.572}{0.083}} \\
\addlinespace[0.05pt]

\midrule
\rowcolor{modelrow}\multicolumn{17}{c}{\textbf{InternVL3-8B}} \\
\textcolor{fullkv}{\textbf{Full KV}} & \textcolor{fullkv}{Upper} 
& \multicolumn{3}{c}{\textcolor{fullkv}{91.1}} 
& \multicolumn{3}{c}{\textcolor{fullkv}{81.6}} 
& \multicolumn{3}{c}{\textcolor{fullkv}{77.8}} 
& \multicolumn{3}{c}{\textcolor{fullkv}{55.3}} 
& \multicolumn{3}{c}{\textcolor{fullkv}{1.111}} \\
\midrule

\textbf{SnapKV} & Base 
& 89.5 & 85.4 & 75.1 
& 80.7 & 78.8 & 72.3 
& 77.4 & 75.9 & 71.6 
& 55.3 & 55.2 & 55.2 
& 1.083 & 0.999 & 0.804 \\
\rowcolor{mixrow} 
& +\texttt{MixKV} 
& 89.5 & 86.4 & 75.5 
& 81.2 & 79.2 & 73.6 
& \best{77.6} & 76.0 & 72.2 
& 55.3 & 55.2 & 55.1 
& \best{1.103} & 1.016 & 0.821 \\
\rowcolor{baconrow} 
& +\texttt{BACON} 
& \best{\gain{90.0}{0.5}} & \best{\gain{88.7}{3.3}} & \best{\gain{84.9}{9.8}} 
& \best{\gain{81.4}{0.7}} & \best{\gain{81.2}{2.4}} & \best{\gain{76.7}{4.4}} 
& \same{77.4}{0.0} & \best{\gain{77.3}{1.4}} & \best{\gain{75.6}{4.0}} 
& \best{\gain{55.4}{0.1}} & \best{\gain{55.3}{0.1}} & \best{\same{55.2}{0.0}} 
& \gain{1.102}{0.019} & \best{\gain{1.020}{0.021}} & \best{\gain{0.856}{0.052}} \\
\addlinespace[0.05pt]

\textbf{PyramidKV} & Base 
& 87.6 & 82.5 & 69.2 
& 78.4 & 76.1 & 68.8 
& 76.4 & 74.8 & 71.1 
& 55.2 & 55.2 & 55.1 
& 0.977 & 0.902 & 0.688 \\
\rowcolor{mixrow} 
& +\texttt{MixKV} 
& 87.7 & 83.4 & 69.4 
& 79.4 & 76.9 & 68.9 
& 76.7 & 75.2 & 71.7 
& 55.3 & 55.3 & 55.2 
& 1.024 & 0.936 & 0.719 \\
\rowcolor{baconrow} 
& +\texttt{BACON} 
& \best{\gain{89.6}{2.0}} & \best{\gain{87.7}{5.2}} & \best{\gain{80.7}{11.5}} 
& \best{\gain{81.1}{2.7}} & \best{\gain{79.7}{3.6}} & \best{\gain{74.4}{5.6}} 
& \best{\gain{77.5}{1.1}} & \best{\gain{76.9}{2.1}} & \best{\gain{74.3}{3.2}} 
& \best{\gain{55.3}{0.1}} & \best{\gain{55.4}{0.2}} & \best{\gain{55.3}{0.2}} 
& \best{\gain{1.027}{0.050}} & \best{\gain{0.945}{0.043}} & \best{\gain{0.782}{0.094}} \\
\addlinespace[0.05pt]

\textbf{AdaKV} & Base 
& 89.5 & 85.7 & 76.9 
& 80.7 & 78.5 & 72.4 
& \best{77.2} & 75.5 & 72.6 
& 55.6 & 55.2 & 55.2 
& 1.092 & 0.983 & 0.847 \\
\rowcolor{mixrow} 
& +\texttt{MixKV} 
& 89.3 & 86.4 & 77.2 
& 81.1 & 79.3 & 73.2 
& 76.9 & 76.1 & 72.5 
& 55.3 & 55.2 & 55.1 
& 1.091 & 1.027 & 0.869 \\
\rowcolor{baconrow} 
& +\texttt{BACON} 
& \best{\same{89.5}{0.0}} & \best{\gain{88.6}{2.9}} & \best{\gain{85.8}{8.9}} 
& \best{\gain{81.4}{0.7}} & \best{\gain{80.4}{1.9}} & \best{\gain{76.8}{4.4}} 
& \drop{77.1}{0.1} & \best{\gain{76.8}{1.3}} & \best{\gain{75.4}{2.8}} 
& \best{\same{55.6}{0.0}} & \best{\same{55.2}{0.0}} & \best{\gain{55.3}{0.1}} 
& \best{\gain{1.102}{0.010}} & \best{\gain{1.030}{0.047}} & \best{\gain{0.872}{0.025}} \\
\addlinespace[0.05pt]

\midrule
\rowcolor{modelrow}\multicolumn{17}{c}{\textbf{Qwen3-VL-30B-A3B}} \\
\textcolor{fullkv}{\textbf{Full KV}} & \textcolor{fullkv}{Upper} 
& \multicolumn{3}{c}{\textcolor{fullkv}{95.5}} 
& \multicolumn{3}{c}{\textcolor{fullkv}{84.3}} 
& \multicolumn{3}{c}{\textcolor{fullkv}{74.4}} 
& \multicolumn{3}{c}{\textcolor{fullkv}{52.56}} 
& \multicolumn{3}{c}{\textcolor{fullkv}{0.328}} \\
\midrule

\textbf{SnapKV} & Base 
& 94.7 & 91.5 & 78.9 
& 83.4 & 81.8 & 75.8 
& 74.7 & 73.2 & 70.4 
& 52.33 & 52.67 & 51.56 
& 0.354 & 0.348 & 0.273 \\
\rowcolor{mixrow} 
& +\texttt{MixKV} 
& 95.3 & 92.7 & 81.0 
& \best{84.0} & 83.0 & 78.8 
& 74.9 & 74.8 & 72.4 
& 52.78 & 53.00 & 51.89 
& \best{0.357} & 0.426 & 0.334 \\
\rowcolor{baconrow} 
& +\texttt{BACON} 
& \best{\gain{95.4}{0.7}} & \best{\gain{95.1}{3.6}} & \best{\gain{90.3}{11.4}} 
& \gain{83.8}{0.4} & \best{\gain{83.7}{1.9}} & \best{\gain{81.4}{5.6}} 
& \best{\gain{75.1}{0.4}} & \best{\gain{74.9}{1.7}} & \best{\gain{73.6}{3.2}} 
& \best{\gain{53.11}{0.78}} & \best{\gain{53.00}{0.33}} & \best{\gain{52.33}{0.77}} 
& \drop{0.347}{0.007} & \best{\gain{0.435}{0.087}} & \best{\gain{0.355}{0.082}} \\
\addlinespace[0.05pt]

\textbf{PyramidKV} & Base 
& 86.9 & 88.0 & 74.4 
& 80.3 & 80.0 & 73.2 
& 72.3 & 72.6 & 68.5 
& 52.89 & 52.22 & 51.78 
& 0.326 & 0.338 & 0.266 \\
\rowcolor{mixrow} 
& +\texttt{MixKV} 
& 89.3 & 90.2 & 76.3 
& 82.3 & 82.9 & 77.2 
& 74.0 & 74.5 & 71.8 
& 53.44 & 52.89 & \best{52.11} 
& 0.401 & \best{0.398} & 0.370 \\
\rowcolor{baconrow} 
& +\texttt{BACON} 
& \best{\gain{94.8}{7.9}} & \best{\gain{94.3}{6.3}} & \best{\gain{87.2}{12.8}} 
& \best{\gain{83.7}{3.4}} & \best{\gain{83.1}{3.1}} & \best{\gain{79.3}{6.1}} 
& \best{\gain{74.7}{2.4}} & \best{\gain{74.6}{2.0}} & \best{\gain{73.2}{4.7}} 
& \best{\gain{53.44}{0.55}} & \best{\gain{53.56}{1.34}} & \gain{52.00}{0.22} 
& \best{\gain{0.425}{0.099}} & \gain{0.375}{0.037} & \best{\gain{0.387}{0.121}} \\
\addlinespace[0.05pt]

\textbf{AdaKV} & Base 
& 94.8 & 91.8 & 80.8 
& 83.0 & 81.4 & 75.2 
& 74.9 & 73.2 & 70.8 
& 52.44 & 52.33 & 52.44 
& 0.347 & 0.325 & 0.293 \\
\rowcolor{mixrow} 
& +\texttt{MixKV} 
& 95.5 & 93.5 & 83.3 
& \best{84.4} & 82.8 & 78.5 
& 74.8 & \best{74.7} & 72.9 
& 52.89 & 52.56 & 51.78 
& 0.351 & 0.414 & 0.353 \\
\rowcolor{baconrow} 
& +\texttt{BACON} 
& \best{\gain{95.7}{0.9}} & \best{\gain{95.0}{3.2}} & \best{\gain{90.6}{9.8}} 
& \gain{83.3}{0.3} & \best{\gain{83.7}{2.3}} & \best{\gain{80.7}{5.5}} 
& \best{\gain{75.0}{0.1}} & \gain{74.5}{1.3} & \best{\gain{73.2}{2.4}} 
& \best{\gain{53.11}{0.67}} & \best{\gain{53.00}{0.67}} & \best{\gain{53.67}{1.23}} 
& \best{\gain{0.352}{0.005}} & \best{\gain{0.417}{0.092}} & \best{\gain{0.361}{0.068}} \\

\bottomrule
\end{tabular}
\end{adjustbox}
}
\vspace{-3mm}
\end{table*}

\noindent \textbf{Structurally calibrated boundary evidence.}
BACON combines the original boundary estimate with its intra-layer coherence and inter-layer persistence:
\begin{equation}
V^{l,h}_i
=
E^{l,h}_i
+
L^{l,h}_i
+
T^{l,h}_i.
\label{eq:structurally_calibrated_evidence}
\end{equation}
In vector form,
\begin{equation}
\mathbf{v}^{l,h}
=
(\mathbf{I}+\mathbf{P}_r)\mathbf{e}^{l,h}
+
\mathbf{t}^{l,h},
\label{eq:structural_calibration_form}
\end{equation}
where $\mathbf{t}^{l,h}$ denotes the inter-layer persistence vector. BACON thus converts noisy last-query saliency into structurally calibrated boundary evidence without task-specific weighting.

% Required packages: booktabs, xcolor[table], adjustbox
% Highlight color
\definecolor{baconrow}{RGB}{255,245,245}
\definecolor{mixrow}{RGB}{248,248,248}
\definecolor{fullkv}{RGB}{112,112,112}

\begin{table*}[t]
\centering
\caption{\textbf{Results on \bestvideo{video understanding benchmarks} with Qwen2-VL-7B.} }
\label{tab:video_results}
\vspace{-3mm}
{\fontsize{5pt}{5.2pt}\selectfont
\setlength{\tabcolsep}{2.15pt}
\renewcommand{\arraystretch}{0.7}
\setlength{\aboverulesep}{0.85pt}
\setlength{\belowrulesep}{0.85pt}
\setlength{\cmidrulesep}{0.85pt}
\setlength{\heavyrulewidth}{0.45pt}
\setlength{\lightrulewidth}{0.25pt}
\setlength{\cmidrulewidth}{0.25pt}
\begin{adjustbox}{width=\textwidth}
\begin{tabular}{@{}ll*{15}{c}@{}}
\toprule
\multicolumn{2}{c}{}
& \multicolumn{12}{c}{\textbf{VATEX}}
& \multicolumn{3}{c}{\textbf{NextQA}} \\
\cmidrule(lr){3-14}\cmidrule(lr){15-17}
\textbf{Method} & \textbf{Variant}
& \multicolumn{3}{c}{\textbf{CIDEr}}
& \multicolumn{3}{c}{\textbf{BLEU-4}}
& \multicolumn{3}{c}{\textbf{METEOR}}
& \multicolumn{3}{c}{\textbf{ROUGE-L}}
& \multicolumn{3}{c}{\textbf{WUPS}} \\
\cmidrule(lr){3-5}\cmidrule(lr){6-8}\cmidrule(lr){9-11}\cmidrule(lr){12-14}\cmidrule(lr){15-17}
& & \textbf{512} & \textbf{256} & \textbf{128}
& \textbf{512} & \textbf{256} & \textbf{128}
& \textbf{512} & \textbf{256} & \textbf{128}
& \textbf{512} & \textbf{256} & \textbf{128}
& \textbf{512} & \textbf{256} & \textbf{128} \\
\midrule
\textcolor{fullkv}{\textbf{Full KV}} & \textcolor{fullkv}{Upper}
& \multicolumn{3}{c}{\textcolor{fullkv}{49.61}}
& \multicolumn{3}{c}{\textcolor{fullkv}{21.91}}
& \multicolumn{3}{c}{\textcolor{fullkv}{23.83}}
& \multicolumn{3}{c}{\textcolor{fullkv}{44.15}}
& \multicolumn{3}{c}{\textcolor{fullkv}{26.30}} \\
\midrule

\textbf{SnapKV} & Base
& 48.14 & 46.51 & \bestvideo{46.27}
& 21.22 & 20.60 & 20.26
& 23.70 & 23.22 & 22.18
& 43.80 & 43.53 & 42.68
& 25.97 & 25.69 & 25.84 \\
\rowcolor{mixrow} & +\texttt{MixKV}
& 48.55 & 47.68 & 45.95
& 21.34 & 20.82 & 20.32
& 23.00 & 23.42 & 22.07
& 43.90 & 43.55 & \bestvideo{42.93}
& 25.99 & 25.93 & 25.60 \\
\rowcolor{baconrow} & +\texttt{BACON}
& \bestvideo{48.67} & \bestvideo{48.02} & 46.02
& \bestvideo{21.38} & \bestvideo{21.26} & \bestvideo{20.38}
& \bestvideo{23.79} & \bestvideo{23.42} & \bestvideo{22.63}
& \bestvideo{43.96} & \bestvideo{43.57} & 42.79
& \bestvideo{26.15} & \bestvideo{25.94} & \bestvideo{26.02} \\
\addlinespace[0.5pt]

\textbf{AdaKV} & Base
& 48.20 & 46.20 & 45.37
& 21.55 & 20.43 & 20.28
& 23.62 & 23.29 & 22.85
& 43.85 & 43.26 & 42.67
& 25.72 & 25.70 & 25.65 \\
\rowcolor{mixrow} & +\texttt{MixKV}
& 48.68 & 47.46 & \bestvideo{45.39}
& 21.22 & 20.78 & 20.34
& 23.68 & 23.28 & 22.87
& 43.92 & 43.32 & 42.68
& \bestvideo{25.99} & \bestvideo{26.01} & 25.86 \\
\rowcolor{baconrow} & +\texttt{BACON}
& \bestvideo{48.82} & \bestvideo{47.85} & 45.33
& \bestvideo{21.55} & \bestvideo{21.24} & \bestvideo{20.54}
& \bestvideo{23.68} & \bestvideo{23.45} & \bestvideo{23.15}
& \bestvideo{44.06} & \bestvideo{43.41} & \bestvideo{42.76}
& 25.96 & 25.95 & \bestvideo{25.86} \\
\addlinespace[0.5pt]

\textbf{PyramidKV} & Base
& 46.09 & 46.03 & 44.36
& 20.58 & \bestvideo{20.29} & 19.26
& \bestvideo{23.28} & 22.86 & 22.02
& 43.25 & 42.41 & 42.18
& 25.88 & 25.69 & 25.63 \\
\rowcolor{mixrow} & +\texttt{MixKV}
& 46.13 & 45.67 & 44.50
& 20.48 & 20.29 & 19.55
& 23.22 & 22.87 & 22.11
& 43.30 & \bestvideo{42.69} & 42.16
& 25.87 & 25.72 & 25.52 \\
\rowcolor{baconrow} & +\texttt{BACON}
& \bestvideo{46.57} & \bestvideo{46.22} & \bestvideo{45.70}
& \bestvideo{20.81} & 20.25 & \bestvideo{19.70}
& 23.26 & \bestvideo{22.89} & \bestvideo{22.24}
& \bestvideo{43.31} & 42.65 & \bestvideo{42.36}
& \bestvideo{26.04} & \bestvideo{25.80} & \bestvideo{25.65} \\
\addlinespace[0.5pt]

\textbf{SparseMM} & Base
& 47.91 & 47.16 & 46.12
& 20.79 & 20.24 & 19.77
& 23.52 & 23.18 & 22.24
& 43.67 & 43.53 & 42.63
& \bestvideo{26.37} & 25.94 & 25.71 \\
\rowcolor{mixrow} & +\texttt{MixKV}
& 48.65 & 47.44 & 45.99
& 20.99 & 20.54 & 19.82
& 23.46 & 23.10 & \bestvideo{22.55}
& 43.85 & 43.22 & 42.66
& 26.13 & 25.98 & 25.83 \\
\rowcolor{baconrow} & +\texttt{BACON}
& \bestvideo{48.90} & \bestvideo{47.57} & \bestvideo{46.38}
& \bestvideo{21.28} & \bestvideo{20.81} & \bestvideo{20.03}
& \bestvideo{23.68} & \bestvideo{23.31} & 22.38
& \bestvideo{44.06} & \bestvideo{43.56} & \bestvideo{42.74}
& 26.33 & \bestvideo{26.17} & \bestvideo{26.07} \\
\bottomrule
\end{tabular}
\end{adjustbox}
}
\vspace{-5mm}
\end{table*}
% Required packages:
% \usepackage{booktabs}
% \usepackage{multirow}
% \usepackage[table]{xcolor}
% \usepackage{graphicx}

\definecolor{grey}{rgb}{0.5, 0.5, 0.5}
\definecolor{mixrow}{RGB}{246,246,246}
\definecolor{baconrow}{RGB}{235,241,248}

\begin{table*}[t]
\centering
\caption{\textbf{Results on \bestgui{ScreenSpot GUI grounding benchmark} with Qwen2-VL-7B.}
}
\vspace{-3mm}
\label{tab:qwen2vl_gui}
{\fontsize{3pt}{1.2pt}\selectfont
\setlength{\tabcolsep}{2.2pt}
\makebox[\textwidth][c]{%
\resizebox{1.02\textwidth}{!}{%
\renewcommand{\arraystretch}{0.8}
\setlength{\aboverulesep}{0.85pt}
\setlength{\belowrulesep}{0.85pt}
\setlength{\cmidrulesep}{0.85pt}
\setlength{\heavyrulewidth}{0.45pt}
\setlength{\lightrulewidth}{0.25pt}
\setlength{\cmidrulewidth}{0.25pt}
\begin{tabular}{l l cc cc cc cc cc cc cc}
\toprule
\multirow{2}{*}{\textbf{Method}} 
& \multirow{2}{*}{\textbf{Variant}}
& \multicolumn{2}{c}{\textbf{Mobile Text}}
& \multicolumn{2}{c}{\textbf{Mobile Icon}}
& \multicolumn{2}{c}{\textbf{Desktop Text}}
& \multicolumn{2}{c}{\textbf{Desktop Icon}}
& \multicolumn{2}{c}{\textbf{Web Text}}
& \multicolumn{2}{c}{\textbf{Web Icon}}
& \multicolumn{2}{c}{\textbf{Average}} \\
\cmidrule(lr){3-4}
\cmidrule(lr){5-6}
\cmidrule(lr){7-8}
\cmidrule(lr){9-10}
\cmidrule(lr){11-12}
\cmidrule(lr){13-14}
\cmidrule(lr){15-16}
& & \textbf{128} & \textbf{64}
  & \textbf{128} & \textbf{64}
  & \textbf{128} & \textbf{64}
  & \textbf{128} & \textbf{64}
  & \textbf{128} & \textbf{64}
  & \textbf{128} & \textbf{64}
  & \textbf{128} & \textbf{64} \\
\midrule

\textcolor{grey}{\textbf{Full KV}} 
& \textcolor{grey}{--}
& \multicolumn{2}{c}{\textcolor{grey}{39.4}}
& \multicolumn{2}{c}{\textcolor{grey}{32.1}}
& \multicolumn{2}{c}{\textcolor{grey}{17.5}}
& \multicolumn{2}{c}{\textcolor{grey}{10.0}}
& \multicolumn{2}{c}{\textcolor{grey}{7.0}}
& \multicolumn{2}{c}{\textcolor{grey}{6.8}}
& \multicolumn{2}{c}{\textcolor{grey}{18.8}} \\
\midrule

\textbf{SnapKV}
& Base
& 24.5 & 15.4 
& 10.5 & 4.8 
& 19.1 & 9.3 
& 4.3 & 4.3 
& 5.7 & 6.5 
& 6.8 & 5.8 
& 11.8 & 7.7 \\
\rowcolor{mixrow}
& + \texttt{MixKV}
& 24.9 & 15.8 
& \bestgui{10.8} & 4.8 
& 19.6 & 10.8 
& 4.3 & 4.3 
& 6.1 & \bestgui{6.5} 
& 6.8 & 6.3 
& 12.1 & 8.1 \\
\rowcolor{baconrow}
& + \texttt{BACON}
& \bestgui{24.9} & \bestgui{15.8} 
& 10.7 & \bestgui{5.7} 
& \bestgui{21.1} & \bestgui{13.7} 
& \bestgui{4.3} & \bestgui{5.7} 
& \bestgui{7.0} & 6.4 
& \bestgui{8.2} & \bestgui{6.3} 
& \bestgui{12.7} & \bestgui{9.0} \\
\midrule

\textbf{AdaKV}
& Base
& 26.1 & 15.4 
& 11.8 & 4.2 
& 19.1 & 11.3 
& 5.7 & 5.0 
& 4.8 & 6.5 
& 6.3 & 5.8 
& 12.3 & 8.0 \\
\rowcolor{mixrow}
& + \texttt{MixKV}
& 26.6 & 14.6 
& 11.3 & 4.8 
& \bestgui{20.1} & 11.3 
& 5.7 & 5.0 
& 5.7 & 5.7 
& 6.3 & 6.3 
& 12.6 & 8.0 \\
\rowcolor{baconrow}
& + \texttt{BACON}
& \bestgui{26.6} & \bestgui{15.4} 
& \bestgui{12.0} & \bestgui{5.0} 
& 19.5 & \bestgui{13.9} 
& \bestgui{5.7} & \bestgui{5.0} 
& \bestgui{5.7} & \bestgui{6.7} 
& \bestgui{6.8} & \bestgui{6.3} 
& \bestgui{12.7} & \bestgui{8.7} \\
\midrule

\textbf{PyramidKV}
& Base
& 24.5 & \bestgui{13.9} 
& 8.2 & 6.6 
& 20.1 & 9.8 
& 3.6 & 5.0 
& 6.1 & 6.1 
& 6.2 & 5.3 
& 11.5 & 7.8 \\
\rowcolor{mixrow}
& + \texttt{MixKV}
& 24.5 & 13.2 
& 8.7 & \bestgui{7.0} 
& 19.1 & 10.3 
& 3.6 & 5.0 
& 6.5 & 6.1 
& \bestgui{6.6} & 5.8 
& 11.5 & 7.9 \\
\rowcolor{baconrow}
& + \texttt{BACON}
& \bestgui{24.8} & 13.5 
& \bestgui{8.9} & 6.7 
& \bestgui{20.1} & \bestgui{10.3} 
& \bestgui{4.3} & \bestgui{5.7} 
& \bestgui{7.8} & \bestgui{6.2} 
& 6.3 & \bestgui{5.8} 
& \bestgui{12.0} & \bestgui{8.0} \\
\midrule

\textbf{SparseMM}
& Base
& 22.7 & 15.8 
& 9.2 & 4.4 
& 18.6 & 9.8 
& 4.6 & 4.3 
& 7.0 & 4.5 
& \bestgui{8.7} & 5.8 
& 11.9 & 7.4 \\
\rowcolor{mixrow}
& + \texttt{MixKV}
& 21.5 & 14.6 
& 10.2 & 4.4 
& 17.0 & 10.3 
& 2.9 & 4.3 
& 7.4 & 3.9 
& 8.2 & 5.8 
& 11.2 & 7.2 \\
\rowcolor{baconrow}
& + \texttt{BACON}
& \bestgui{22.7} & \bestgui{16.9} 
& \bestgui{10.2} & \bestgui{4.5} 
& \bestgui{19.6} & \bestgui{10.3} 
& \bestgui{4.9} & \bestgui{5.0} 
& \bestgui{7.6} & \bestgui{5.2} 
& 8.3 & \bestgui{8.2} 
& \bestgui{12.2} & \bestgui{8.4} \\

\bottomrule
\end{tabular}
}%
}
}
\vspace{-2.5mm}
\end{table*}
% Required packages:
% \usepackage{booktabs}
% \usepackage{multirow}
% \usepackage[table]{xcolor}
% \usepackage{graphicx}
% \usepackage{rotating}

\definecolor{grey}{rgb}{0.5, 0.5, 0.5}
\definecolor{mixrow}{RGB}{246,246,246}
\definecolor{baconrow}{RGB}{255,253,242}

% Adjustable table style
\newcommand{\LongBenchFontSize}{5.8pt}
\newcommand{\LongBenchFontLeading}{6.2pt}
\newcommand{\LongBenchRowStretch}{0.7}
\newcommand{\LongBenchColSep}{2.2pt}

% Adjustable rotated dataset-header style
\newcommand{\LongBenchHeaderFontSize}{4pt}
\newcommand{\LongBenchHeaderFontLeading}{2pt}
\newcommand{\lbheader}[1]{%
  \rotatebox[origin=c]{45}{%
    \fontsize{\LongBenchHeaderFontSize}{\LongBenchHeaderFontLeading}\selectfont\textbf{#1}%
  }%
}

\begin{table*}[t]
\centering
\caption{\textbf{Results on \bestlb{LongBench long context understanding benchmark} with Mistral-7B-Instruct-v0.2.}
}
\vspace{-3mm}
\label{tab:longbench_mistral}
{\fontsize{\LongBenchFontSize}{\LongBenchFontLeading}\selectfont
\setlength{\tabcolsep}{\LongBenchColSep}
\resizebox{\textwidth}{!}{%
\renewcommand{\arraystretch}{\LongBenchRowStretch}
\begin{tabular}{lccccccccccccccccc}
\toprule
\multirow{2}{*}{\textbf{Methods}}
& \multicolumn{3}{c}{\textbf{Single-Doc QA}}
& \multicolumn{3}{c}{\textbf{Multi-Doc QA}}
& \multicolumn{3}{c}{\textbf{Summarization}}
& \multicolumn{3}{c}{\textbf{Few-shot}}
& \multicolumn{2}{c}{\textbf{Synthetic}}
& \multicolumn{2}{c}{\textbf{Code}}
& \multirow{2}{*}{\textbf{Avg.}} \\
\cmidrule(lr){2-4}
\cmidrule(lr){5-7}
\cmidrule(lr){8-10}
\cmidrule(lr){11-13}
\cmidrule(lr){14-15}
\cmidrule(lr){16-17}
& \lbheader{NrtvQA}
& \lbheader{Qasper}
& \lbheader{MF-en}
& \lbheader{HotpotQA}
& \lbheader{2WikiMQA}
& \lbheader{Musique}
& \lbheader{GovReport}
& \lbheader{QMSum}
& \lbheader{MultiNews}
& \lbheader{TREC}
& \lbheader{TriviaQA}
& \lbheader{SAMSum}
& \lbheader{PCount}
& \lbheader{PRe}
& \lbheader{Lcc}
& \lbheader{RB-P}
& \\
\midrule

\textcolor{grey}{\textbf{Full KV}}
& \textcolor{grey}{26.77} & \textcolor{grey}{32.51} & \textcolor{grey}{49.36}
& \textcolor{grey}{43.58} & \textcolor{grey}{27.35} & \textcolor{grey}{18.86}
& \textcolor{grey}{33.09} & \textcolor{grey}{24.38} & \textcolor{grey}{27.03}
& \textcolor{grey}{71.00} & \textcolor{grey}{86.23} & \textcolor{grey}{42.99}
& \textcolor{grey}{2.89} & \textcolor{grey}{86.98}
& \textcolor{grey}{47.16} & \textcolor{grey}{48.03}
& \textcolor{grey}{41.76} \\

\midrule
\multicolumn{18}{c}{\textbf{KV Cache Budget = 1024}} \\
\midrule

\textbf{SnapKV}
& 25.06 & 28.92 & 49.17
& \bestlb{40.41} & 26.11 & 18.14
& 25.91 & 23.99 & 25.73
& 67.00 & 86.24 & 41.57
& 2.98 & \bestlb{87.48}
& 46.13 & 46.09
& 40.06 \\
\rowcolor{mixrow}
+ \texttt{MixKV}
& 24.73 & 29.88 & 48.62
& 39.68 & 26.41 & 18.56
& 26.15 & 23.70 & 26.16
& 69.00 & 86.25 & 43.25
& 3.21 & 85.84
& \bestlb{46.27} & 46.32
& 40.25 \\
\rowcolor{baconrow}
+ \texttt{BACON}
& \bestlb{25.89} & \bestlb{31.15} & \bestlb{49.42}
& 40.27 & \bestlb{26.47} & \bestlb{18.94}
& \bestlb{26.32} & \bestlb{24.38} & \bestlb{26.37}
& \bestlb{71.00} & \bestlb{86.43} & \bestlb{43.56}
& \bestlb{3.36} & 87.21
& 46.19 & \bestlb{46.68}
& \bestlb{40.85} \\

\textbf{AdaKV}
& 25.28 & 31.02 & 48.53
& 41.06 & 26.49 & 19.28
& 25.74 & 24.02 & 25.55
& 68.50 & 86.28 & 42.43
& 2.94 & 88.10
& 46.35 & 46.58
& 40.51 \\
\rowcolor{mixrow}
+ \texttt{MixKV}
& 25.60 & 30.94 & 48.79
& 40.72 & 26.56 & 19.25
& \bestlb{26.74} & 23.98 & 25.68
& 69.00 & 86.25 & \bestlb{43.39}
& 2.96 & 86.88
& 46.78 & 46.01
& 40.60 \\
\rowcolor{baconrow}
+ \texttt{BACON}
& \bestlb{26.99} & \bestlb{31.11} & \bestlb{48.90}
& \bestlb{41.43} & \bestlb{26.68} & \bestlb{19.87}
& 26.65 & \bestlb{24.57} & \bestlb{25.91}
& \bestlb{71.00} & \bestlb{86.49} & 43.33
& \bestlb{3.03} & \bestlb{88.23}
& \bestlb{46.94} & \bestlb{46.64}
& \bestlb{41.11} \\

\textbf{PyramidKV}
& 24.68 & 27.34 & 48.53
& 39.91 & 25.78 & 18.71
& 25.73 & 23.50 & 25.37
& 68.50 & 85.80 & 41.29
& 2.66 & \bestlb{86.73}
& 45.60 & 46.30
& 39.78 \\
\rowcolor{mixrow}
+ \texttt{MixKV}
& 23.72 & 29.85 & 48.12
& 39.17 & \bestlb{26.78} & 19.29
& 26.67 & 23.59 & 26.47
& 70.00 & 85.63 & \bestlb{43.00}
& 3.13 & 84.05
& 46.10 & 45.50
& 40.07 \\
\rowcolor{baconrow}
+ \texttt{BACON}
& \bestlb{24.86} & \bestlb{30.94} & \bestlb{48.74}
& \bestlb{40.51} & 26.73 & \bestlb{19.52}
& \bestlb{26.92} & \bestlb{23.67} & \bestlb{27.48}
& \bestlb{71.00} & \bestlb{86.14} & 42.98
& \bestlb{3.21} & 86.28
& \bestlb{46.22} & \bestlb{46.63}
& \bestlb{40.74} \\

\midrule
\multicolumn{18}{c}{\textbf{KV Cache Budget = 512}} \\
\midrule

\textbf{SnapKV}
& 24.15 & 26.67 & 48.94
& 37.38 & 25.90 & 17.18
& 23.80 & 22.83 & 24.32
& 65.50 & 85.88 & 42.07
& 3.17 & 87.13
& 45.01 & 45.80
& 39.11 \\
\rowcolor{mixrow}
+ \texttt{MixKV}
& 24.04 & 27.03 & 48.18
& 37.79 & 26.03 & 17.54
& 24.56 & 23.43 & \bestlb{25.20}
& 66.50 & 86.16 & 42.92
& 3.26 & 87.04
& 45.39 & 45.82
& 39.43 \\
\rowcolor{baconrow}
+ \texttt{BACON}
& \bestlb{25.21} & \bestlb{27.81} & \bestlb{49.17}
& \bestlb{38.47} & \bestlb{26.30} & \bestlb{17.74}
& \bestlb{24.83} & \bestlb{23.95} & 24.83
& \bestlb{69.00} & \bestlb{86.36} & \bestlb{42.92}
& \bestlb{3.44} & \bestlb{87.20}
& \bestlb{45.85} & \bestlb{45.96}
& \bestlb{39.94} \\

\textbf{AdaKV}
& 24.67 & 26.93 & 48.39
& 38.47 & \bestlb{26.07} & 17.14
& 23.80 & 23.50 & 24.10
& 66.00 & 86.11 & 42.00
& \bestlb{3.29} & 87.33
& 45.70 & \bestlb{46.02}
& 39.35 \\
\rowcolor{mixrow}
+ \texttt{MixKV}
& 24.30 & 27.85 & 48.40
& 38.33 & 25.76 & \bestlb{18.34}
& 24.39 & 23.61 & 24.80
& 67.50 & 85.92 & 42.49
& 2.99 & 87.12
& \bestlb{46.53} & 45.77
& 39.63 \\
\rowcolor{baconrow}
+ \texttt{BACON}
& \bestlb{24.85} & \bestlb{28.50} & \bestlb{48.84}
& \bestlb{38.49} & 25.83 & 18.30
& \bestlb{24.83} & \bestlb{23.62} & \bestlb{25.34}
& \bestlb{70.00} & \bestlb{86.35} & \bestlb{43.30}
& 3.14 & \bestlb{87.64}
& 45.96 & 45.81
& \bestlb{40.05} \\

\textbf{PyramidKV}
& \bestlb{23.39} & 24.80 & 47.56
& 38.23 & 25.29 & 17.32
& 23.47 & 23.02 & 23.67
& 66.00 & 85.31 & 41.48
& 2.86 & 86.30
& 45.35 & 43.54
& 38.60 \\
\rowcolor{mixrow}
+ \texttt{MixKV}
& 22.99 & 25.38 & 47.89
& 38.21 & 24.09 & 17.76
& 23.98 & 22.78 & \bestlb{24.92}
& 67.50 & 85.41 & 41.73
& 3.19 & 86.41
& 45.24 & 44.35
& 38.86 \\
\rowcolor{baconrow}
+ \texttt{BACON}
& 23.06 & \bestlb{27.29} & \bestlb{48.13}
& \bestlb{38.41} & \bestlb{26.25} & \bestlb{18.21}
& \bestlb{24.06} & \bestlb{23.06} & 24.76
& \bestlb{69.50} & \bestlb{86.07} & \bestlb{42.33}
& \bestlb{3.22} & \bestlb{86.91}
& \bestlb{45.82} & \bestlb{44.45}
& \bestlb{39.47} \\

\bottomrule
\end{tabular}
}%
}
\vspace{-5mm}
\end{table*}

\subsection{Variance-Constrained Score Calibration}
\label{subsec:variance_calibration}

The structurally calibrated boundary evidence $V^{l,h}_i$ is used as a controlled calibration term rather than a replacement for the backbone score. To ensure comparability across layers and heads, BACON matches the variance of boundary evidence to the local variation of the observation window score.

For each layer-head pair, let $\mathcal{I}_{l,h}$ be the candidate set. We define the scale-matching region under the backbone score as
\begin{equation}
M_{l,h}
=
\min(\rho K_{l,h},|\mathcal{I}_{l,h}|),
\label{eq:scale_matching_size}
\end{equation}
\begin{equation}
\mathcal{M}_{l,h}
=
\operatorname{TopK}_{i\in\mathcal{I}_{l,h}}
\left(B^{l,h}_i, M_{l,h}\right),
\label{eq:scale_matching_set}
\end{equation}
where $\rho$ is an expansion factor, set to $\rho=2$ by default. BACON obtains the calibration coefficient by solving
\begin{equation}
\begin{aligned}
\lambda^{l,h}
=
\arg\min_{\lambda\geq0}
\Big(
&\operatorname{Std}_{i\in\mathcal{M}_{l,h}}
(\lambda V^{l,h}_i) \\
&-
\gamma
\operatorname{Std}_{i\in\mathcal{M}_{l,h}}
(B^{l,h}_i)
\Big)^2 ,
\end{aligned}
\label{eq:lambda_objective}
\end{equation}
where $\gamma$ controls the global calibration strength. This objective has the closed-form solution
\begin{equation}
\lambda^{l,h}
=
\gamma
\cdot
\frac{
\operatorname{Std}_{i\in\mathcal{M}_{l,h}}
(B^{l,h}_i)
}{
\operatorname{Std}_{i\in\mathcal{M}_{l,h}}
(V^{l,h}_i)
+
\epsilon
},
\label{eq:lambda_solution}
\end{equation}
where $\epsilon$ is a small constant for numerical stability. The final BACON score is
\begin{equation}
S^{l,h}_i
=
B^{l,h}_i
+
\lambda^{l,h}V^{l,h}_i.
\label{eq:bacon_score}
\end{equation}
This preserves the stable observation window score while injecting boundary evidence at a calibrated scale.

Finally, BACON selects the retained token set
\begin{equation}
\mathcal{K}_{l,h}
=
\operatorname{TopK}_{i\in\mathcal{I}_{l,h}}
\left(
S^{l,h}_i,
K_{l,h}
\right),
\label{eq:topk_selection}
\end{equation}
and constructs the compressed cache by retaining $\{\mathbf{K}^{l}_{h,i},\mathbf{V}^{l}_{h,i}\mid i\in\mathcal{K}_{l,h}\}$. We further provide a theoretical justification of BACON in \ref{append:theory}.

\section{Experiment}

\subsection{Experiment Settings}

\noindent \textbf{Model Details.}
We evaluate BACON across diverse model architectures and scales. For multimodal understanding, we use LLaVA-NeXT-Mistral-7B \citep{liu2024llavanext}, InternVL3-8B \citep{zhu2025internvl3}, Qwen2-VL-7B-Instruct \citep{wang2024qwen2}, and Qwen3-VL-30B-A3B-Instruct \citep{bai2025qwen3}. For text-only evaluation, we use Mistral-7B-Instruct-v0.2 \citep{jiang2023mistral7b} and Llama3.1-8B-Instruct \citep{grattafiori2024llama}.

\noindent \textbf{Benchmark Details.}
We evaluate BACON on image understanding, video understanding, GUI grounding, and long-context text tasks. For \best{image understanding}, we use DocVQA \citep{mathew2021docvqa}, TextVQA \citep{singh2019towards}, MMMU \citep{yue2024mmmu}, ChartQA \citep{masry2022chartqa}, and TextCaps \citep{sidorov2020textcaps}. For \bestvideo{video understanding}, we use VATEX \citep{wang2019vatex} and NextQA \citep{xiao2021next}. For \bestgui{GUI grounding}, we use ScreenSpot \citep{li2025screenspot}. For \bestlb{text understanding}, we use LongBench \citep{bai2024longbench}.

% Required packages:
% \usepackage{booktabs}
% \usepackage{multirow}
% \usepackage[table]{xcolor}
% \usepackage{graphicx}
% \usepackage{pifont}

\definecolor{ablationrow}{RGB}{236,236,236}
\definecolor{baconrow}{RGB}{245,245,245}
\definecolor{checkc}{RGB}{45,120,75}
\definecolor{crossc}{RGB}{170,65,65}

\newcommand{\cmark}{\textcolor{checkc}{\ding{51}}}
\newcommand{\xmark}{\textcolor{crossc}{\ding{55}}}

\begin{table*}[t]
\centering
\caption{\textbf{Ablation and efficiency analysis of BACON.}
Left: component ablation on LLaVA-NeXT-Mistral-7B under budget 64.
Right: efficiency comparison with fixed input length 32000.}
\label{tab:ablation_efficiency}
\vspace{-3mm}

\begin{minipage}[t]{0.64\textwidth}
\centering
\tiny
\setlength{\tabcolsep}{2.1pt}
\renewcommand{\arraystretch}{0.54}
\setlength{\aboverulesep}{0.12ex}
\setlength{\belowrulesep}{0.12ex}
\setlength{\cmidrulesep}{0.05ex}

\resizebox{\linewidth}{!}{
\begin{tabular}{lccc ccccc}
\toprule
\multirow{2}{*}{\textbf{Backbone}}
& \multirow{2}{*}{$\boldsymbol{E}$}
& \multirow{2}{*}{$\boldsymbol{L}$}
& \multirow{2}{*}{$\boldsymbol{T}$}
& \multicolumn{5}{c}{\textbf{Budget 64}} \\
\cmidrule(lr){5-9}
& & &
& \textbf{ChartQA} & \textbf{DocVQA} & \textbf{TextVQA} & \textbf{MMMU} & \textbf{TextCaps} \\
\midrule

\multirow{5}{*}{\textbf{SnapKV}}
& \xmark & \xmark & \xmark
& 47.2 & 46.2 & 58.9 & 34.8 & 0.442 \\
& \cmark & \xmark & \cmark
& 47.8 & 52.1 & 62.0 & 34.8 & 0.482 \\
& \cmark & \cmark & \xmark
& 47.9 & 51.7 & 61.6 & 34.8 & 0.465 \\
& \xmark & \cmark & \cmark
& 46.8 & 50.7 & 60.9 & 34.8 & 0.508 \\
\rowcolor{ablationrow}
& \cmark & \cmark & \cmark
& \textbf{49.2} & \textbf{53.9} & \textbf{62.8} & \textbf{34.8} & \textbf{0.512} \\

\midrule

\multirow{5}{*}{\textbf{AdaKV}}
& \xmark & \xmark & \xmark
& 47.4 & 47.6 & 57.8 & 34.8 & 0.441 \\
& \cmark & \xmark & \cmark
& 48.7 & 52.3 & 59.3 & 34.8 & 0.495 \\
& \cmark & \cmark & \xmark
& 47.9 & 54.8 & 60.9 & 34.8 & 0.477 \\
& \xmark & \cmark & \cmark
& 47.7 & 51.4 & 59.5 & 34.8 & 0.469 \\
\rowcolor{ablationrow}
& \cmark & \cmark & \cmark
& \textbf{49.0} & \textbf{55.2} & \textbf{61.5} & \textbf{34.8} & \textbf{0.510} \\

\midrule

\multirow{5}{*}{\textbf{SparseMM}}
& \xmark & \xmark & \xmark
& 49.4 & 57.5 & 65.2 & 34.8 & 0.490 \\
& \cmark & \xmark & \cmark
& 49.6 & 58.1 & 66.2 & 34.7 & 0.553 \\
& \cmark & \cmark & \xmark
& 49.4 & 58.3 & 66.7 & 34.9 & 0.540 \\
& \xmark & \cmark & \cmark
& 49.2 & 57.9 & 66.4 & 34.7 & 0.547 \\
\rowcolor{ablationrow}
& \cmark & \cmark & \cmark
& \textbf{49.8} & \textbf{58.3} & \textbf{67.3} & \textbf{34.9} & \textbf{0.572} \\

\bottomrule
\end{tabular}
}
\end{minipage}
\hfill
\begin{minipage}[t]{0.34\textwidth}
\centering
\tiny
\setlength{\tabcolsep}{3.0pt}
\renewcommand{\arraystretch}{0.88}
\setlength{\aboverulesep}{0.12ex}
\setlength{\belowrulesep}{0.12ex}

\resizebox{\linewidth}{!}{
\begin{tabular}{llcc}
\toprule
\textbf{Backbone} & \textbf{Method} & \textbf{Lat.} & \textbf{Mem.} \\
\midrule
Full KV & Base & 63.93 & 22.27 \\
\midrule
\multirow{3}{*}{SnapKV}
& Base & 28.60 & 18.27 \\
& + \texttt{MixKV} & 28.66 & 18.27 \\
\rowcolor{baconrow}
& + \texttt{BACON} & 28.68 & 18.28 \\
\midrule
\multirow{3}{*}{SparseMM}
& Base & 28.62 & 17.74 \\
& + \texttt{MixKV} & 28.15 & 17.74 \\
\rowcolor{baconrow}
& + \texttt{BACON} & 28.01 & 17.75 \\
\midrule
\multirow{3}{*}{PyramidKV}
& Base & 28.52 & 18.27 \\
& + \texttt{MixKV} & 28.72 & 18.27 \\
\rowcolor{baconrow}
& + \texttt{BACON} & 28.21 & 18.29 \\
\midrule
\multirow{3}{*}{AdaKV}
& Base & 28.67 & 17.74 \\
& + \texttt{MixKV} & 28.40 & 17.74 \\
\rowcolor{baconrow}
& + \texttt{BACON} & 28.34 & 17.76 \\
\bottomrule
\end{tabular}
}
\end{minipage}

\vspace{-5mm}
\end{table*}

\noindent \textbf{Implementation Details.}
We apply BACON to various KV cache compression methods, including SnapKV, PyramidKV, AdaKV, and SparseMM, under different cache budgets. More experiment details are provided in \ref{append:exp}.

\subsection{Main Results}

\noindent \textbf{Performance on image understanding benchmarks.}
Table~\ref{tab:main_results_compact} summarizes BACON integrated with representative KV compression baselines across multiple MLLMs, benchmarks, and cache budgets. The results show three main advantages.
(i) \textit{Consistent effectiveness.} BACON consistently improves compressed inference, especially under aggressive budgets where sparse visual evidence is more likely to be discarded. For example, on Qwen2-VL-7B with PyramidKV at budget 64, BACON improves DocVQA by +18.7 points and TextVQA by +8.3 points; similar gains with SnapKV and AdaKV indicate that last-query calibration helps recover visual evidence weakened by observation window aggregation.
(ii) \textit{Broad applicability.} BACON benefits diverse compression paradigms, including attention-based selection, adaptive allocation, and head-/layer-wise budget allocation. On LLaVA-NeXT-Mistral-7B at budget 64, it improves PyramidKV by +8.6 points on DocVQA and SparseMM by +2.1 points on TextVQA. Since BACON only refines within-head token scores without changing the compression operator or budget allocation, it can be directly integrated into existing KV compression pipelines.
(iii) \textit{Model compatibility.} BACON generalizes from 7B-scale MLLMs to stronger architectures, including InternVL3-8B and the 30B-scale MoE model Qwen3-VL-30B-A3B. With PyramidKV at budget 64, it improves Qwen3-VL-30B-A3B by +12.8 points on DocVQA and +6.1 points on TextVQA, supporting its role as a plug-and-play calibration method across model families and scales.

\noindent \textbf{Performance on video understanding benchmarks.}
Beyond static image benchmarks, we evaluate BACON on video understanding tasks, where relevant evidence can be distributed across frames. As shown in Table~\ref{tab:video_results}, BACON improves compressed inference on both VATEX and NextQA across different compression backbones and cache budgets. On VATEX, BACON yields broad gains across methods and budgets, suggesting better preservation of fine-grained visual-temporal cues. On NextQA, it improves WUPS in most settings, indicating that the calibrated retention signal also benefits video question answering. These results show that BACON generalizes beyond image-level grounding while remaining compatible with diverse compression strategies. A few metric-level fluctuations suggest that some videos may benefit from broader temporal coverage retained by the base score. Overall, BACON provides a complementary calibration signal for compressed video-language inference.

\noindent \textbf{Performance on GUI grounding benchmarks.}
GUI grounding requires models to identify fine-grained interface elements, making it sensitive to information loss under KV compression. We evaluate BACON on ScreenSpot with Qwen2-VL-7B to test whether it preserves interface evidence under compressed caches. As shown in Table~\ref{tab:qwen2vl_gui}, BACON improves average performance across different compression backbones under both 128 and 64 budgets, with stronger gains under the tighter budget. This suggests that BACON helps retain weak but critical GUI evidence that may be diluted by observation window aggregation. Minor drops on a few subsets reflect the sensitivity of GUI grounding to small retention changes under tight budgets, rather than a systematic limitation. These results validate BACON as a complementary calibration signal for memory-efficient GUI agent deployment.

\noindent \textbf{Performance on long-context text benchmarks.}
To evaluate BACON beyond multimodal tasks, we further test it on LongBench with Mistral-7B-Instruct-v0.2. As shown in Table~\ref{tab:longbench_mistral}, BACON improves average performance across compression backbones and cache budgets, showing that boundary-aware calibration is not limited to visual tokens and can also refine token scoring in language-only settings. The rare regressions are mainly concentrated in Synthetic and Code tasks, where performance often depends on exact positions, copied strings, or short-range syntactic dependencies. In these cases, small changes in retained tokens can affect task-specific exactness, making boundary calibration less uniformly beneficial than in semantic understanding tasks. Nevertheless, the overall gains show that BACON remains effective in text-only settings, while its larger improvements on MLLM benchmarks further support our motivation that sparse multimodal evidence is particularly vulnerable to observation window aggregation. Additional results on Llama3.1-8B-Instruct are provided in~\ref{appendix:llama}, and more evaluations under extreme multimodal scenarios~\citep{wang2025divide,fu2025video} are provided in~\ref{append:extreme}.

\subsection{Ablation Study \& Analysis}

\noindent \textbf{Ablation of BACON components.}
Table~\ref{tab:ablation_efficiency} ablates the three key components of BACON, including boundary evidence $E$, intra-layer coherence $L$, and inter-layer persistence $T$, under budget 64.
Across different compression backbones, the full BACON consistently achieves the best performance, demonstrating the complementarity of these components under aggressive KV compression.
The gains are especially pronounced on DocVQA, TextVQA, and TextCaps, indicating that OCR-intensive tasks are highly sensitive to the loss of sparse visual evidence caused by observation window aggregation.
Here, $E$ provides a boundary-aware correction for recovering evidence weakened by window aggregation, while $L$ and $T$ strengthen intra-layer coherence and inter-layer persistence to reduce answer-irrelevant high-attention responses.
The weaker performance of partial variants further suggests that effective token retention requires both boundary-sensitive evidence discovery and structure-aware evidence calibration. Additional ablation results on Qwen2-VL-7B and cache budgets are provided in \ref{append:ablation}.
% =========================================================
% Put these in the preamble
% =========================================================
% \usepackage{booktabs}
% \usepackage[table]{xcolor}
% \usepackage{tabularx}
% \usepackage{array}

\definecolor{sensrow}{RGB}{245,245,245}

% Centered X column for tabularx
\newcolumntype{Y}{>{\centering\arraybackslash}X}

% =========================================================
% Sensitivity to m
% =========================================================
\begin{table}[t]
\centering
\caption{\textbf{Sensitivity to the inter-layer persistence range $m$}
on SnapKV under a KV budget of 64.}
\label{tab:sensitivity_m}
\vspace{-1.5mm}

\fontsize{8.5}{9.5}\selectfont
\renewcommand{\arraystretch}{1.02}
\setlength{\aboverulesep}{0.25ex}
\setlength{\belowrulesep}{0.25ex}

\begin{tabularx}{0.89\columnwidth}{YYYY}
\toprule
\textbf{$m$} &
\textbf{DocVQA} &
\textbf{ChartQA} &
\textbf{TextVQA} \\
\midrule
0  & 51.7 & 47.9 & 61.6 \\
2  & 53.4 & 48.2 & 62.1 \\
\rowcolor{sensrow}
\textbf{4$^\ast$} &
\textbf{53.9} &
\textbf{49.2} &
62.8 \\
6  & 53.9 & 48.9 & \textbf{62.9} \\
8  & 53.5 & 48.8 & 62.7 \\
\bottomrule
\end{tabularx}

\vspace{-1.5mm}
\end{table}

% =========================================================
% Sensitivity to r
% =========================================================
\begin{table}[t]
\centering
\caption{\textbf{Sensitivity to the intra-layer neighborhood $r$}
on SnapKV under a KV budget of 64.}
\label{tab:sensitivity_r}
\vspace{-1.5mm}

\fontsize{8.5}{9.5}\selectfont
\renewcommand{\arraystretch}{1.02}
\setlength{\aboverulesep}{0.25ex}
\setlength{\belowrulesep}{0.25ex}

\begin{tabularx}{0.89\columnwidth}{YYYY}
\toprule
\textbf{$r$} &
\textbf{DocVQA} &
\textbf{ChartQA} &
\textbf{TextVQA} \\
\midrule
0 & 52.1 & 47.8 & 62.0 \\
3 & 53.7 & 48.8 & 62.8 \\
\rowcolor{sensrow}
\textbf{5$^\ast$} &
\textbf{53.9} &
\textbf{49.2} &
62.8 \\
7 & 53.7 & 49.1 & \textbf{62.9} \\
9 & 53.7 & 48.9 & 62.5 \\
\bottomrule
\end{tabularx}

\vspace{-4.5mm}
\end{table}
\noindent \textbf{Efficiency Analysis of BACON.}
Table~\ref{tab:ablation_efficiency} evaluates the inference latency and peak memory consumption of BACON when integrated with representative KV cache compression backbones. BACON preserves the efficiency benefits of compressed inference and introduces negligible additional runtime or memory overhead over the original compression methods. This is because BACON only refines token-retention scores during prefill and does not change the compressed cache size or decoding procedure. Therefore, its accuracy gains are obtained without additional inference cost.

\noindent \textbf{Hyperparameter Sensitivity Analysis of BACON.}
Tables~\ref{tab:sensitivity_m} and~\ref{tab:sensitivity_r} examine the sensitivity of BACON to the inter-layer persistence range $m$ and intra-layer neighborhood radius $r$, respectively, under SnapKV with a KV-cache budget of 64. Overall, BACON exhibits consistent performance over a broad range of both hyperparameters, suggesting that its effectiveness is not tied to a particular parameter choice. More importantly, disabling either cross-layer persistence or local neighborhood aggregation generally leads to inferior performance, supporting the role of these two components in reinforcing reliable evidence while suppressing isolated attention noise. Performance remains stable across moderate ranges and only deteriorates mildly when the aggregation scope becomes excessively large, likely because increasingly distant signals introduce less relevant evidence. These results demonstrate that BACON is robust to its hyperparameter configuration, while the default setting ($m=4$, $r=5$) offers a practical balance between evidence reinforcement and noise control.

Additional robustness evaluations under alternative prompt formulations and two-stage chain-of-thought inference are provided in~\ref{append:prompt}, where BACON remains effective across different prompt and inference configurations.
\section{Conclusion}
\label{sec:conclusion}
In this work, we analyze attention-based KV cache compression in MLLMs and identify a key limitation of observation window attention: its aggregation can dilute sparse visual evidence under tight budgets despite providing stable token-importance estimates.
Motivated by this insight, we propose BACON, a plug-and-play boundary attention calibration method that refines observation window retention scores with last-query evidence.
BACON further suppresses noisy boundary saliency through intra-layer coherence and inter-layer persistence, thereby producing a stronger retention signal without modifying the original compression pipeline.
Extensive experiments spanning multimodal understanding, video reasoning, GUI grounding, and long-context text tasks across different model scales, architectures, compression methods, and cache budgets show that BACON consistently improves existing KV compression methods while requiring no extra method modification or additional inference cost. 

\section{Limitations}
BACON is designed as a training-free and plug-and-play mechanism, enabling broad compatibility with existing KV cache compression pipelines without additional model updates. This design choice naturally focuses BACON on improving token-retention quality under a fixed compression policy. As a result, it does not explicitly exploit training signals to adapt retention behavior to sample-level evidence distributions, task characteristics, or input complexity. Incorporating BACON into training could further enable the model to learn more adaptive compression strategies for different samples and tasks. In addition, KV cache compression methods integrated with BACON can be applied to many latency-sensitive scenarios, such as real-time multimodal assistants, GUI agents, embodied systems, and edge-device deployment. 
However, their use should be carefully controlled in high-precision or safety-critical applications. If compression is applied too aggressively or without appropriate validation, answer-critical visual or textual evidence may be discarded, leading to degraded model performance and potentially affecting downstream functionality. 
Future deployment should therefore consider task-specific reliability requirements and provide safeguards, such as budget constraints, confidence checks, or fallback mechanisms to less compressed inference when necessary.

% =========================================================
% Bibliography

\bibliography{custom}

% =========================================================
% Appendix
% =========================================================
\appendix

\section{Theoretical Justification of BACON}
\label{append:theory}

We provide a theoretical interpretation of two properties underlying BACON:
(i) answer-relevant evidence tends to emerge at the last query before generation, and
(ii) genuine evidence exhibits stronger coherence across tokens and persistence across layers than incidental attention noise.

\paragraph{Boundary-emergent evidence.}
Let $D$ denote the last input position before answer generation. At layer $l$, its attention output is

\begin{equation}
\begin{aligned}
s_i^{(l)}
&=
\frac{(q_D^{(l)})^\top k_i^{(l)}}{\sqrt{d}}, \\
\alpha_i^{(l)}
&=
\operatorname{softmax}(s^{(l)})_i, \\
o_D^{(l)}
&=
\sum_{i \leq D}
\alpha_i^{(l)} v_i^{(l)}.
\end{aligned}
\end{equation}

The first answer token is predicted directly from the final hidden state at this position,

\begin{equation}
p(y_1 \mid \mathbf{X})
=
\operatorname{softmax}
\left(
W_{\mathrm{out}} h_D^{(l_{\max})}
\right).
\end{equation}

Thus, the boundary position serves as the immediate interface between multimodal context encoding and answer generation: it has causal access to the complete input prefix and directly determines the next-token prediction.

This property is also reflected in the optimization objective. Let

\begin{equation}
g^{(l)}
=
\frac{\partial \mathcal{L}_1}{\partial o_D^{(l)}},
\qquad
u_i^{(l)}
=
-
(g^{(l)})^\top v_i^{(l)},
\end{equation}

where $u_i^{(l)}$ denotes the first-order predictive utility of token $i$. A gradient-descent update to its attention logit satisfies

\begin{equation}
\Delta s_i^{(l)}
=
\eta \alpha_i^{(l)}
\left(
u_i^{(l)}-\bar{u}^{(l)}
\right),
\end{equation}

where $\bar{u}^{(l)}=\sum_j\alpha_j^{(l)}u_j^{(l)}$.
Therefore, tokens contributing above-average utility to answer prediction are explicitly encouraged to receive larger attention from the last query. This explains the stronger emergence of answer-relevant evidence at the generation boundary, consistent with the empirical observation in Fig.~2(b).

\paragraph{Structured evidence versus noise.}
Let $z_{i,l}$ denote the centered boundary-attention response of token $i$ at layer $l$. We model an answer-relevant token as

\begin{equation}
z_{i,l}
=
a_i b_l+\epsilon_{i,l},
\qquad i\in\mathcal{S},
\end{equation}

where $a_i$ captures the token-level evidence strength, $b_l$ captures its propagation across layers, and $\epsilon_{i,l}$ denotes perturbation. For unrelated tokens,

\begin{equation}
z_{i,l}
=
\epsilon_{i,l},
\qquad
\mathbb{E}[\epsilon_{i,l}] \approx 0.
\end{equation}

Hence, over the token--layer response matrix,

\begin{equation}
Z_{\mathcal{S}}
=
ab^\top + \Xi,
\qquad
Z_{\mathrm{noise}}
\approx \Xi.
\end{equation}

Genuine evidence therefore contains a structured component shared across tokens and layers, whereas incidental responses are substantially less coherent.

\paragraph{Intra-layer coherence.}
For a local neighborhood $\mathcal{N}_r(i)$, define

\begin{equation}
C_{i,l}
=
\frac{1}{|\mathcal{N}_r(i)|}
\sum_{j\in\mathcal{N}_r(i)}
z_{j,l}.
\end{equation}

For an evidence region,

\begin{equation}
\mathbb{E}[C_{i,l}\mid i\in\mathcal{S}]
=
b_l
\frac{1}{|\mathcal{N}_r(i)|}
\sum_{j\in\mathcal{N}_r(i)} a_j,
\end{equation}

which remains strong when neighboring tokens share the same evidence factor. In contrast, for centered weakly correlated noise,

\begin{equation}
\begin{aligned}
\mathbb{E}[C_{i,l}\mid\mathrm{noise}]
&\approx 0, \\
\operatorname{Var}(C_{i,l}\mid\mathrm{noise})
&\propto
\frac{1}{|\mathcal{N}_r(i)|}.
\end{aligned}
\end{equation}

Local aggregation therefore preserves spatially or semantically coherent evidence while suppressing isolated attention spikes, motivating BACON's intra-layer coherence term.

\paragraph{Inter-layer persistence.}
Similarly, define the response aggregated over the most recent $m$ layers as

\begin{equation}
P_{i,l}^{(m)}
=
\frac{1}{m}
\sum_{t=0}^{m-1}
z_{i,l-t}.
\end{equation}

For an evidence token,

\begin{equation}
\mathbb{E}
\left[
P_{i,l}^{(m)}
\mid
i\in\mathcal{S}
\right]
=
a_i
\frac{1}{m}
\sum_{t=0}^{m-1}
b_{l-t},
\end{equation}

which remains nonzero when the same evidence is repeatedly propagated through successive layers. For weakly correlated noise,

\begin{equation}
\begin{aligned}
\mathbb{E}
\left[
P_{i,l}^{(m)}
\mid
\mathrm{noise}
\right]
&\approx 0, \\
\operatorname{Var}
\left(
P_{i,l}^{(m)}
\mid
\mathrm{noise}
\right)
&\propto
\frac{1}{m_{\mathrm{eff}}}.
\end{aligned}
\end{equation}

Cross-layer aggregation therefore suppresses transient activations while retaining evidence that remains consistently salient across depth, motivating BACON's inter-layer persistence term.

In conclusion, last-query attention provides an answer-conditioned evidence signal, while intra-layer coherence and inter-layer persistence validate this signal along the token and layer dimensions. BACON thereby favors evidence that is both locally coherent and persistently propagated while suppressing isolated and transient noise.

\section{Detailed Experiment Setting}
\label{append:exp}
\subsection{Implementation Details.}
All experiments were conducted on NVIDIA RTX 3090 and NVIDIA H800 GPUs.
BACON is implemented as a plug-and-play modification to existing KV cache compression methods and is applied only during the prefill-stage cache selection.
It does not require model retraining, additional supervision, or changes to the original model parameters.
For each compression method, we keep its original layer-wise and head-wise cache budget allocation unchanged, and only modify the within-head token retention score.
Unless otherwise specified, all results are obtained under the same decoding settings as the corresponding baseline.
\subsection{Model Details.}
We introduce more details on the models we used in our experiments:

\textbf{LLaVA-NeXT-Mistral-7B.}
LLaVA-NeXT-Mistral-7B is a representative open-source MLLM built upon the Mistral-7B language backbone. It follows the common vision-language architecture that combines a pretrained visual encoder, a multimodal projector, and an autoregressive LLM. Compared with earlier LLaVA models, LLaVA-NeXT improves high-resolution image understanding, OCR ability, and visual reasoning by using dynamic high-resolution image processing and stronger instruction-tuning data. In our experiments, this model serves as a strong 7B-scale MLLM baseline for evaluating KV cache compression under image-text reasoning tasks.

\textbf{Qwen2-VL-7B-Instruct.}
Qwen2-VL-7B-Instruct is an instruction-tuned vision-language model from the Qwen2-VL series. It supports flexible visual inputs, including images with different resolutions and aspect ratios, and is designed for fine-grained visual understanding, document comprehension, OCR-related reasoning, and general multimodal dialogue. The model adopts a unified multimodal generation framework in which visual tokens are injected into the language model and processed together with textual instructions. We use Qwen2-VL-7B-Instruct as a high-performing 7B-scale MLLM to evaluate whether BACON remains effective on modern multimodal architectures with stronger visual perception capability.

\textbf{Qwen3-VL-30B-A3B-Instruct.}
Qwen3-VL-30B-A3B-Instruct is a larger-scale multimodal model from the Qwen3-VL family. It is designed for comprehensive vision-language understanding, including image understanding, video understanding, spatial reasoning, grounding, and long-form visual comprehension. Unlike dense 7B-scale MLLMs, this model adopts a larger mixture-of-experts-style architecture, with around 30B total parameters and a smaller number of active parameters during inference. We include Qwen3-VL-30B-A3B-Instruct to examine whether BACON can generalize beyond 7B-scale dense MLLMs and remain effective for larger multimodal backbones with stronger reasoning capacity.

\textbf{InternVL3-8B.}
InternVL3-8B is an advanced open-source MLLM from the InternVL3 series. Different from many MLLMs that mainly adapt a pretrained text-only LLM to visual inputs, InternVL3 emphasizes native multimodal pretraining, where multimodal and linguistic abilities are jointly acquired during training. The model is designed to improve multimodal perception, reasoning, GUI understanding, document analysis, and other vision-language tasks. We use InternVL3-8B as another 8B-scale MLLM backbone to test the robustness of BACON across different multimodal training paradigms and architectural designs.

\textbf{Mistral-7B-Instruct-v0.2.}
Mistral-7B-Instruct-v0.2 is an instruction-tuned text-only LLM based on the Mistral-7B architecture. It is a compact yet strong decoder-only transformer model and has been widely used in long-context language understanding and generation benchmarks. Since it does not contain a visual encoder or multimodal projector, all input tokens are textual. We include this model to evaluate whether the proposed boundary-aware KV retention strategy is also beneficial for text-only long-context inference, rather than being limited to multimodal visual-token compression.

\textbf{Llama-3.1-8B-Instruct.}
Llama-3.1-8B-Instruct is an instruction-tuned text-only LLM from the Llama 3.1 family. It is an autoregressive decoder-only transformer optimized for multilingual dialogue and general text generation. The model uses grouped-query attention to improve inference scalability and supports long-context processing. In our experiments, Llama-3.1-8B-Instruct provides an additional text-only backbone for evaluating BACON on long-context language tasks, allowing us to test whether the proposed retention mechanism generalizes across different LLM families.

\subsection{Benchmark Details.}
We provide details for each benchmark used in our experiments:

\textbf{DocVQA.}
DocVQA evaluates visual question answering over document images. Given a document image and a natural-language question, the model is required to locate relevant textual or structural evidence from the document and generate the correct answer. This benchmark emphasizes OCR ability, document layout understanding, and fine-grained evidence localization. We report ANLS as the evaluation metric.

\textbf{ChartQA.}
ChartQA evaluates question answering over chart images. It requires the model to understand chart structures, recognize visual elements such as bars, lines, legends, axes, and labels, and perform numerical or logical reasoning based on the chart content. We report relaxed accuracy, which allows minor numerical deviations from the reference answer.

\textbf{TextVQA.}
TextVQA evaluates question answering over natural images containing scene text. The model must jointly recognize visual content, read text appearing in the image, and reason over both visual and textual evidence to answer the question. This benchmark is useful for measuring whether compressed KV caches preserve sparse but answer-critical OCR evidence.

\textbf{TextCaps.}
TextCaps evaluates image captioning with reading comprehension. Unlike standard image captioning benchmarks, many correct captions require the model to incorporate text appearing in the image. Therefore, TextCaps tests both visual description ability and scene-text understanding.

\textbf{MMMU.}
MMMU is a multi-discipline multimodal reasoning benchmark built from college-level materials such as exams, quizzes, and textbooks. It covers diverse subjects including science, engineering, medicine, business, and the humanities. Solving MMMU requires models to combine visual perception, domain knowledge, and multi-step reasoning. We use the testmini split in our experiments.

\textbf{VATEX.}
VATEX is a large-scale video captioning benchmark. Given a video clip, the model is required to generate a natural-language description of the visual content. Since useful evidence is distributed across multiple frames, VATEX is used to evaluate the effectiveness of BACON under longer multimodal contexts. We report standard captioning metrics, including CIDEr, BLEU-4, METEOR, and ROUGE-L.

\textbf{NExT-QA.}
NExT-QA is a video question answering benchmark designed to evaluate temporal and causal reasoning. The questions often require understanding action sequences, object interactions, temporal order, and causal relations across video frames. This benchmark allows us to test whether BACON can preserve temporally distributed evidence in video-language reasoning.

\textbf{ScreenSpot.}
ScreenSpot evaluates GUI grounding ability. Given a screenshot and a natural-language instruction, the model must identify the target user-interface element. The benchmark covers mobile, web, and desktop environments, and includes both text-based and icon/widget-based elements. This setting is challenging because the target evidence is often small, spatially localized, and visually similar to nearby interface components.

\textbf{LongBench.}
LongBench evaluates long-context understanding for text-only LLMs. It includes tasks from single-document question answering, multi-document question answering, summarization, few-shot learning, synthetic reasoning, and code completion. In our experiments, LongBench is used to assess whether BACON also generalizes to text-only long-context inference, where the input consists entirely of textual tokens rather than visual tokens.

\textbf{HR-Bench.}
HR-Bench is a high-resolution visual understanding benchmark designed to evaluate fine-grained perception in 4K and 8K images. It covers both single-instance and cross-instance perception, requiring models to recognize small visual details and integrate information across spatially separated regions. This benchmark is particularly suitable for evaluating BACON under long visual contexts, where critical evidence may occupy only a small fraction of the high-resolution image and can be easily discarded during aggressive KV-cache compression.

\textbf{Video-MME.}
Video-MME is a comprehensive video understanding benchmark covering diverse video domains and varying temporal durations. Given a video and a question, the model is required to reason over visual information distributed across multiple frames to produce the correct answer. The benchmark evaluates a broad range of video understanding abilities, including perception, temporal reasoning, and information integration. Its long and temporally distributed visual contexts allow us to assess whether BACON can retain sparse but important evidence during video KV cache compression.

\subsection{Baseline Details.}
We provide more details on the baseline used to compare with our methods in the experiment:

\textbf{SnapKV.}
SnapKV is a training-free KV cache compression method that exploits the observation that attention patterns during prefilling can reveal tokens that are likely to remain important during decoding. It estimates token importance using attention from a local observation window near the end of the prompt and retains the most salient KV pairs under a given budget. SnapKV provides a simple and efficient attention-based compression baseline.

\textbf{PyramidKV.}
PyramidKV compresses the KV cache with a pyramid-like budget allocation strategy across transformer layers. Instead of assigning the same retention budget to every layer, it allocates different budgets according to layer depth, reflecting the intuition that different layers contribute differently to context retention and generation. This baseline evaluates whether BACON can improve token selection under non-uniform layer-wise budget allocation.
% Required packages:
% \usepackage{booktabs}
% \usepackage{multirow}
% \usepackage[table]{xcolor}
% \usepackage{graphicx}
% \usepackage{rotating}

\definecolor{grey}{rgb}{0.5, 0.5, 0.5}
\definecolor{mixrow}{RGB}{246,246,246}
\definecolor{baconrow}{RGB}{255,253,242}

% Adjustable table style
\newcommand{\LlamaLongBenchFontSize}{5.8pt}
\newcommand{\LlamaLongBenchFontLeading}{6.2pt}
\newcommand{\LlamaLongBenchRowStretch}{0.7}
\newcommand{\LlamaLongBenchColSep}{2.2pt}

% Adjustable rotated dataset-header style
\newcommand{\LlamaLongBenchHeaderFontSize}{4pt}
\newcommand{\LlamaLongBenchHeaderFontLeading}{2pt}
\newcommand{\llbheader}[1]{%
  \rotatebox[origin=c]{45}{%
    \fontsize{\LlamaLongBenchHeaderFontSize}{\LlamaLongBenchHeaderFontLeading}\selectfont\textbf{#1}%
  }%
}

\begin{table*}[t]
\centering
\caption{\textbf{Results on \bestlb{LongBench long context understanding benchmark} with Llama3.1-8B-Instruct.}
}
\label{tab:longbench_llama}
{\fontsize{\LlamaLongBenchFontSize}{\LlamaLongBenchFontLeading}\selectfont
\setlength{\tabcolsep}{\LlamaLongBenchColSep}
\resizebox{\textwidth}{!}{%
\renewcommand{\arraystretch}{\LlamaLongBenchRowStretch}
\begin{tabular}{lccccccccccccccccc}
\toprule
\multirow{2}{*}{\textbf{Methods}}
& \multicolumn{6}{c}{\textbf{Information Localization}}
& \multicolumn{6}{c}{\textbf{Information Aggregation}}
& \multicolumn{4}{c}{\textbf{Synthetic / Code}}
& \multirow{2}{*}{\textbf{Avg.}} \\
\cmidrule(lr){2-7}
\cmidrule(lr){8-13}
\cmidrule(lr){14-17}
& \llbheader{NrtvQA}
& \llbheader{Qasper}
& \llbheader{MF-en}
& \llbheader{HotpotQA}
& \llbheader{2WikiMQA}
& \llbheader{Musique}
& \llbheader{GovReport}
& \llbheader{QMSum}
& \llbheader{MultiNews}
& \llbheader{TREC}
& \llbheader{TriviaQA}
& \llbheader{SAMSum}
& \llbheader{PCount}
& \llbheader{PRe}
& \llbheader{Lcc}
& \llbheader{RB-P}
& \\
\midrule

\textcolor{grey}{\textbf{Full KV}}
& \textcolor{grey}{28.29} & \textcolor{grey}{45.53} & \textcolor{grey}{54.94}
& \textcolor{grey}{56.02} & \textcolor{grey}{46.66} & \textcolor{grey}{31.34}
& \textcolor{grey}{35.19} & \textcolor{grey}{25.28} & \textcolor{grey}{27.16}
& \textcolor{grey}{72.50} & \textcolor{grey}{91.65} & \textcolor{grey}{43.59}
& \textcolor{grey}{8.91} & \textcolor{grey}{99.50}
& \textcolor{grey}{52.73} & \textcolor{grey}{49.20}
& \textcolor{grey}{48.03} \\

\midrule
\multicolumn{18}{c}{\textbf{KV Cache Budget = 1024}} \\
\midrule

\textbf{SnapKV}
& 27.54 & 44.04 & 54.64
& 55.24 & 46.18 & 30.70
& 28.30 & 24.42 & 25.78
& 68.00 & 91.83 & 42.44
& 8.46 & 99.50
& 52.27 & 48.72
& 46.75 \\
\rowcolor{mixrow}
+ \texttt{MixKV}
& 27.61 & 44.06 & 54.47
& 55.66 & 45.76 & 31.11
& 28.48 & 24.59 & 26.41
& 69.50 & 91.74 & 42.58
& 7.76 & 99.50
& \bestlb{52.88} & 48.64
& 46.92 \\
\rowcolor{baconrow}
+ \texttt{BACON}
& \bestlb{27.83} & \bestlb{44.31} & \bestlb{54.81}
& \bestlb{56.02} & \bestlb{46.32} & \bestlb{31.56}
& \bestlb{28.72} & \bestlb{24.66} & \bestlb{26.59}
& \bestlb{72.00} & \bestlb{91.90} & \bestlb{43.07}
& \bestlb{8.59} & \bestlb{99.50}
& 52.51 & \bestlb{48.90}
& \bestlb{47.33} \\

\textbf{AdaKV}
& 28.03 & 43.34 & \bestlb{55.36}
& 55.67 & 45.90 & 31.23
& 28.47 & 24.25 & 25.94
& 68.50 & 91.62 & 42.25
& \bestlb{8.43} & 99.50
& 52.28 & 49.23
& 46.88 \\
\rowcolor{mixrow}
+ \texttt{MixKV}
& \bestlb{28.21} & 43.38 & 54.71
& 55.79 & 46.10 & \bestlb{31.55}
& 29.06 & 24.09 & 26.15
& 70.50 & 91.73 & 43.11
& 8.26 & 99.50
& \bestlb{52.68} & 49.25
& 47.13 \\
\rowcolor{baconrow}
+ \texttt{BACON}
& 28.13 & \bestlb{44.05} & 55.27
& \bestlb{56.04} & \bestlb{46.64} & 31.52
& \bestlb{29.31} & \bestlb{24.28} & \bestlb{26.31}
& \bestlb{72.50} & \bestlb{91.73} & \bestlb{43.63}
& 8.29 & \bestlb{99.50}
& 52.62 & \bestlb{49.34}
& \bestlb{47.45} \\

\textbf{PyramidKV}
& 28.06 & 42.63 & 54.19
& 55.65 & 46.23 & 31.46
& 27.67 & 24.85 & 25.79
& 68.00 & 91.79 & 41.57
& 8.34 & 99.50
& 51.35 & 48.01
& 46.57 \\
\rowcolor{mixrow}
+ \texttt{MixKV}
& 28.54 & \bestlb{43.18} & 54.67
& 55.56 & 46.06 & 30.82
& 28.26 & 24.76 & \bestlb{26.35}
& 69.50 & 91.34 & 43.05
& 8.18 & 99.50
& 51.58 & \bestlb{48.47}
& 46.86 \\
\rowcolor{baconrow}
+ \texttt{BACON}
& \bestlb{28.57} & 42.89 & \bestlb{54.81}
& \bestlb{55.75} & \bestlb{46.38} & \bestlb{31.73}
& \bestlb{28.45} & \bestlb{25.12} & 25.81
& \bestlb{72.50} & \bestlb{92.34} & \bestlb{43.14}
& \bestlb{8.49} & \bestlb{99.50}
& \bestlb{51.63} & 48.38
& \bestlb{47.22} \\

\midrule
\multicolumn{18}{c}{\textbf{KV Cache Budget = 512}} \\
\midrule

\textbf{SnapKV}
& 27.40 & 38.23 & 53.49
& 55.04 & 45.30 & 29.92
& 25.48 & 24.21 & 24.39
& 64.00 & 92.05 & 41.92
& 8.17 & 99.50
& 52.18 & 47.46
& 45.55 \\
\rowcolor{mixrow}
+ \texttt{MixKV}
& 27.46 & 38.58 & 53.18
& 55.66 & 45.47 & 30.48
& 25.63 & 24.47 & 24.97
& 68.00 & 92.04 & \bestlb{43.25}
& 7.67 & 99.50
& 51.86 & 47.52
& 45.98 \\
\rowcolor{baconrow}
+ \texttt{BACON}
& \bestlb{27.62} & \bestlb{38.79} & \bestlb{53.73}
& \bestlb{55.72} & \bestlb{45.66} & \bestlb{31.10}
& \bestlb{25.94} & \bestlb{24.92} & \bestlb{25.25}
& \bestlb{71.50} & \bestlb{92.26} & 43.05
& \bestlb{8.34} & \bestlb{99.50}
& \bestlb{52.22} & \bestlb{47.73}
& \bestlb{46.46} \\

\textbf{AdaKV}
& 27.38 & 40.76 & \bestlb{53.41}
& 55.67 & 45.53 & 29.81
& 25.96 & 24.38 & 24.61
& 66.00 & 92.35 & 42.27
& 8.01 & 99.50
& 52.75 & 48.26
& 46.04 \\
\rowcolor{mixrow}
+ \texttt{MixKV}
& 27.36 & 42.28 & 53.04
& 55.78 & 45.39 & 29.95
& 26.06 & 24.64 & 24.46
& 69.50 & 92.07 & 43.24
& \bestlb{8.09} & 99.50
& 52.72 & 48.44
& 46.41 \\
\rowcolor{baconrow}
+ \texttt{BACON}
& \bestlb{27.62} & \bestlb{42.53} & 53.20
& \bestlb{55.99} & \bestlb{45.68} & \bestlb{30.04}
& \bestlb{26.19} & \bestlb{24.83} & \bestlb{24.84}
& \bestlb{72.00} & \bestlb{92.71} & \bestlb{43.37}
& 8.04 & \bestlb{99.50}
& \bestlb{53.09} & \bestlb{48.59}
& \bestlb{46.76} \\

\textbf{PyramidKV}
& 27.15 & 40.52 & 53.15
& 55.19 & 45.57 & 29.90
& 25.29 & 24.19 & 24.31
& 62.50 & 91.30 & 42.08
& 8.57 & 99.50
& 50.79 & 46.31
& 45.40 \\
\rowcolor{mixrow}
+ \texttt{MixKV}
& 27.11 & 41.12 & 53.34
& 55.40 & 45.08 & 30.45
& \bestlb{25.64} & 24.49 & 24.39
& 67.00 & 91.06 & 42.71
& 8.59 & 99.50
& 50.74 & 46.67
& 45.83 \\
\rowcolor{baconrow}
+ \texttt{BACON}
& \bestlb{27.89} & \bestlb{41.37} & \bestlb{53.67}
& \bestlb{56.07} & \bestlb{46.46} & \bestlb{30.62}
& 25.57 & \bestlb{24.70} & \bestlb{24.42}
& \bestlb{71.50} & \bestlb{91.77} & \bestlb{43.65}
& \bestlb{8.62} & \bestlb{99.50}
& \bestlb{51.26} & \bestlb{46.82}
& \bestlb{46.49} \\

\bottomrule
\end{tabular}
}%
}
\vspace{-3.5mm}
\end{table*}
% =========================================================
% Required packages
% =========================================================
% \usepackage{booktabs}
% \usepackage[table]{xcolor}
% \usepackage{multirow}
% \usepackage{tabularx}
% \usepackage{array}

\definecolor{extremerow}{RGB}{236,236,236}

\newcolumntype{Y}{>{\centering\arraybackslash}X}

% =========================================================
% Extreme Multimodal Scenarios
% =========================================================
\begin{table*}[t]
\centering
\caption{\textbf{Results under extreme multimodal contexts with Qwen2-VL-7B.}}
\label{tab:extreme_scenarios}
\vspace{-2mm}

\small
\renewcommand{\arraystretch}{1.02}
\setlength{\aboverulesep}{0.18ex}
\setlength{\belowrulesep}{0.18ex}
\setlength{\cmidrulesep}{0.08ex}

\begin{tabularx}{0.85\textwidth}{
    >{\raggedright\arraybackslash}p{2.25cm}
    Y Y
    @{\hspace{8mm}}
    >{\raggedright\arraybackslash}p{2.25cm}
    Y
}
\toprule

\multirow{3}{*}{\textbf{Method}}
& \multicolumn{2}{c}{\textbf{HR-Bench}}
& \multirow{3}{*}{\textbf{Method}}
& \multirow{2}{*}{\textbf{Video-MME}} \\

\cmidrule(lr){2-3}

& \textbf{4K}
& \textbf{8K}
&
& \\

& \textbf{Acc.}
& \textbf{Acc.}
&
& \textbf{Acc.} \\

\midrule

Full KV
& 64.30
& 62.10
&
Full KV
& 53.30 \\

\midrule

\textbf{SnapKV}
& 63.90
& 61.50
&
\textbf{SnapKV}
& 51.70 \\

\quad + MixKV
& 63.60
& 61.50
&
\quad + MixKV
& 51.70 \\

\rowcolor{extremerow}
\quad + \textbf{BACON}
& \textbf{64.30}
& \textbf{61.80}
&
\quad + \textbf{BACON}
& \textbf{52.30} \\

\midrule

\textbf{AdaKV}
& 63.90
& 61.50
&
\textbf{AdaKV}
& 51.70 \\

\quad + MixKV
& 63.90
& 61.80
&
\quad + MixKV
& 51.70 \\

\rowcolor{extremerow}
\quad + \textbf{BACON}
& \textbf{63.90}
& \textbf{62.10}
&
\quad + \textbf{BACON}
& \textbf{52.30} \\

\midrule

\textbf{SparseMM}
& 63.90
& 61.50
&
\textbf{SparseMM}
& 51.70 \\

\quad + MixKV
& 63.90
& 61.80
&
\quad + MixKV
& 52.30 \\

\rowcolor{extremerow}
\quad + \textbf{BACON}
& \textbf{64.30}
& \textbf{62.10}
&
\quad + \textbf{BACON}
& \textbf{53.00} \\

\bottomrule
\end{tabularx}

\vspace{-3mm}
\end{table*}
\textbf{AdaKV.}
AdaKV performs adaptive KV cache compression by dynamically assigning retention budgets according to token or head importance. Compared with fixed-budget strategies, AdaKV aims to preserve more KV pairs in more informative regions while applying stronger compression to less important ones. We include AdaKV to evaluate whether BACON remains effective when the compression budget is adaptively distributed rather than uniformly assigned.
% Required packages:
% \usepackage{booktabs}
% \usepackage{multirow}
% \usepackage[table]{xcolor}
% \usepackage{graphicx}
% \usepackage{pifont}

\definecolor{ablationrow}{RGB}{236,236,236}
\definecolor{checkc}{RGB}{45,120,75}
\definecolor{crossc}{RGB}{170,65,65}

\newcommand{\llavamark}{\textcolor{checkc}{\ding{51}}}
\newcommand{\llavaxmark}{\textcolor{crossc}{\ding{55}}}

\begin{table*}[t]
\centering
\caption{\textbf{Additional ablation results of BACON on LLaVA-NeXT-Mistral-7B under budget 128.}}
\label{tab:appendix_ablation_llava_b128}
\vspace{-2mm}

{\fontsize{5pt}{5.2pt}\selectfont
\setlength{\tabcolsep}{1.7pt}
\renewcommand{\arraystretch}{0.48}
\setlength{\aboverulesep}{0.08ex}
\setlength{\belowrulesep}{0.08ex}
\setlength{\cmidrulesep}{0.03ex}

\resizebox{0.82\textwidth}{!}{
\begin{tabular}{lccc ccccc}
\toprule
\multirow{2}{*}{\textbf{Backbone}}
& \multirow{2}{*}{$\boldsymbol{E}$}
& \multirow{2}{*}{$\boldsymbol{L}$}
& \multirow{2}{*}{$\boldsymbol{T}$}
& \multicolumn{5}{c}{\textbf{Budget 128}} \\
\cmidrule(lr){5-9}
& & &
& \textbf{ChartQA} & \textbf{DocVQA} & \textbf{TextVQA} & \textbf{MMMU} & \textbf{TextCaps} \\
\midrule

\multirow{5}{*}{\textbf{SnapKV}}
& \llavaxmark & \llavaxmark & \llavaxmark
& 48.4 & 55.2 & 63.0 & 34.9 & 0.560 \\
& \llavamark & \llavaxmark & \llavamark
& 49.3 & 56.3 & 63.7 & 34.9 & 0.632 \\
& \llavamark & \llavamark & \llavaxmark
& 49.1 & 57.1 & 65.4 & 34.8 & 0.624 \\
& \llavaxmark & \llavamark & \llavamark
& 48.9 & 56.6 & 64.6 & 34.9 & 0.644 \\
\rowcolor{ablationrow}
& \llavamark & \llavamark & \llavamark
& \textbf{49.6} & \textbf{57.8} & \textbf{66.2} & \textbf{34.9} & \textbf{0.676} \\

\midrule

\multirow{5}{*}{\textbf{AdaKV}}
& \llavaxmark & \llavaxmark & \llavaxmark
& 48.8 & 56.1 & 62.7 & 34.9 & 0.568 \\
& \llavamark & \llavaxmark & \llavamark
& 45.0 & 57.6 & 65.1 & 34.9 & 0.606 \\
& \llavamark & \llavamark & \llavaxmark
& 44.5 & 57.4 & 65.4 & 34.7 & 0.626 \\
& \llavaxmark & \llavamark & \llavamark
& 45.3 & 57.0 & 63.2 & 34.9 & 0.619 \\
\rowcolor{ablationrow}
& \llavamark & \llavamark & \llavamark
& \textbf{50.2} & \textbf{58.4} & \textbf{65.5} & \textbf{34.9} & \textbf{0.641} \\

\midrule

\multirow{5}{*}{\textbf{SparseMM}}
& \llavaxmark & \llavaxmark & \llavaxmark
& 49.8 & 58.9 & 67.4 & 34.7 & 0.600 \\
& \llavamark & \llavaxmark & \llavamark
& 50.2 & 59.2 & \textbf{67.8} & 34.7 & 0.597 \\
& \llavamark & \llavamark & \llavaxmark
& 50.0 & 58.9 & 67.5 & 34.7 & 0.604 \\
& \llavaxmark & \llavamark & \llavamark
& 49.9 & 58.6 & 66.7 & 34.7 & 0.605 \\
\rowcolor{ablationrow}
& \llavamark & \llavamark & \llavamark
& \textbf{50.4} & \textbf{59.3} & \textbf{67.8} & \textbf{34.8} & \textbf{0.626} \\

\bottomrule
\end{tabular}
}
}
\vspace{-3mm}
\end{table*}
% Required packages:
% \usepackage{booktabs}
% \usepackage{multirow}
% \usepackage[table]{xcolor}
% \usepackage{graphicx}
% \usepackage{pifont}

\definecolor{ablationrow}{RGB}{236,236,236}
\definecolor{checkc}{RGB}{45,120,75}
\definecolor{crossc}{RGB}{170,65,65}

\newcommand{\qwencmark}{\textcolor{checkc}{\ding{51}}}
\newcommand{\qwenxmark}{\textcolor{crossc}{\ding{55}}}

\begin{table*}[t]
\centering
\caption{\textbf{Additional ablation results of BACON on Qwen2-VL-7B under budgets 64 and 128.}}
\label{tab:appendix_ablation_qwen}
\vspace{-2mm}

\tiny
\setlength{\tabcolsep}{1.8pt}
\renewcommand{\arraystretch}{0.54}
\setlength{\aboverulesep}{0.12ex}
\setlength{\belowrulesep}{0.12ex}
\setlength{\cmidrulesep}{0.05ex}

\resizebox{\textwidth}{!}{
\begin{tabular}{lccc ccccc ccccc}
\toprule
\multirow{2}{*}{\textbf{Backbone}}
& \multirow{2}{*}{$\boldsymbol{E}$}
& \multirow{2}{*}{$\boldsymbol{L}$}
& \multirow{2}{*}{$\boldsymbol{T}$}
& \multicolumn{5}{c}{\textbf{Budget 64}}
& \multicolumn{5}{c}{\textbf{Budget 128}} \\
\cmidrule(lr){5-9}\cmidrule(lr){10-14}
& & &
& \textbf{ChartQA} & \textbf{DocVQA} & \textbf{TextVQA} & \textbf{MMMU} & \textbf{TextCaps}
& \textbf{ChartQA} & \textbf{DocVQA} & \textbf{TextVQA} & \textbf{MMMU} & \textbf{TextCaps} \\
\midrule

\multirow{5}{*}{\textbf{SnapKV}}
& \qwenxmark & \qwenxmark & \qwenxmark
& 66.2 & 70.1 & 70.3 & 49.6 & 0.787
& 69.6 & 82.1 & 77.0 & 49.8 & 1.141 \\
& \qwencmark & \qwenxmark & \qwencmark
& 68.1 & 84.0 & 77.1 & 50.0 & 1.114
& \textbf{70.4} & 90.6 & 81.9 & 49.8 & 1.410 \\
& \qwencmark & \qwencmark & \qwenxmark
& 68.2 & 83.5 & 76.4 & 49.9 & 1.151
& 69.7 & 88.3 & 81.4 & \textbf{50.0} & 1.401 \\
& \qwenxmark & \qwencmark & \qwencmark
& 67.5 & 84.7 & 77.3 & 50.0 & 1.133
& 69.9 & 87.7 & 81.8 & 49.8 & 1.408 \\
\rowcolor{ablationrow}
& \qwencmark & \qwencmark & \qwencmark
& \textbf{69.6} & \textbf{85.5} & \textbf{78.2} & \textbf{50.0} & \textbf{1.178}
& 70.2 & \textbf{91.5} & \textbf{82.6} & 49.9 & \textbf{1.426} \\

\midrule

\multirow{5}{*}{\textbf{AdaKV}}
& \qwenxmark & \qwenxmark & \qwenxmark
& 66.6 & 69.4 & 70.8 & 49.6 & 0.771
& 69.6 & 81.3 & 75.9 & 49.7 & 1.099 \\
& \qwencmark & \qwenxmark & \qwencmark
& 67.8 & 85.6 & 77.0 & 49.8 & 1.103
& 70.1 & 88.6 & 79.2 & 49.8 & 1.359 \\
& \qwencmark & \qwencmark & \qwenxmark
& 67.9 & 84.3 & 76.4 & 49.8 & \textbf{1.146}
& 69.4 & 85.3 & 80.4 & 49.7 & 1.363 \\
& \qwenxmark & \qwencmark & \qwencmark
& 68.3 & 81.7 & 77.6 & 49.6 & 1.111
& 69.8 & 86.5 & 81.1 & 49.7 & 1.348 \\
\rowcolor{ablationrow}
& \qwencmark & \qwencmark & \qwencmark
& \textbf{69.6} & \textbf{86.2} & \textbf{78.5} & \textbf{49.8} & 1.144
& \textbf{70.2} & \textbf{91.1} & \textbf{81.2} & \textbf{49.8} & \textbf{1.371} \\

\midrule

\multirow{5}{*}{\textbf{SparseMM}}
& \qwenxmark & \qwenxmark & \qwenxmark
& 69.6 & 87.3 & 76.9 & 49.6 & 1.044
& 70.0 & 91.4 & 82.1 & 49.8 & 1.427 \\
& \qwencmark & \qwenxmark & \qwencmark
& 69.9 & 90.5 & 81.1 & 49.8 & 1.388
& \textbf{70.4} & 92.3 & 82.3 & 49.6 & 1.471 \\
& \qwencmark & \qwencmark & \qwenxmark
& 70.0 & 91.9 & 80.7 & 49.8 & 1.415
& 70.1 & 93.0 & 82.1 & 49.8 & 1.503 \\
& \qwenxmark & \qwencmark & \qwencmark
& 69.8 & 89.4 & 81.2 & 49.6 & 1.407
& 69.7 & 92.3 & 82.3 & 49.8 & 1.498 \\
\rowcolor{ablationrow}
& \qwencmark & \qwencmark & \qwencmark
& \textbf{70.2} & \textbf{92.0} & \textbf{81.6} & \textbf{49.8} & \textbf{1.432}
& \textbf{70.4} & \textbf{93.2} & \textbf{82.4} & \textbf{49.8} & \textbf{1.511} \\

\bottomrule
\end{tabular}
}
\vspace{-3mm}
\end{table*}
\textbf{SparseMM.}
SparseMM is a multimodal KV cache compression baseline designed for efficient MLLM inference. It exploits sparsity in multimodal attention and applies head-wise KV retention to reduce redundant visual and textual tokens while preserving task-relevant evidence. Since SparseMM provides optimized head-wise budget allocation, it is a strong baseline for evaluating whether BACON can further improve within-head token selection without changing the original budget assignment.

\textbf{MixKV.}
MixKV is a plug-and-play KV cache compression method for large vision-language models. It selects KV pairs by jointly considering token importance and diversity, aiming to preserve both highly attended tokens and semantically diverse contextual information. MixKV further adopts head-wise adaptive mixing to balance these criteria across attention heads. We use MixKV as a strong plug-in baseline and compare BACON against it under the same models, budgets, and evaluation settings.

\subsection{Evaluation Metrics Details.}
We follow the official evaluation protocol of each benchmark. Specifically, DocVQA is evaluated with ANLS, which measures normalized string similarity and accounts for minor OCR or formatting variations in document question answering. ChartQA uses relaxed accuracy, allowing small numerical deviations from the reference answer. TextVQA, MMMU, NExT-QA, ScreenSpot, and LongBench are evaluated with accuracy-based metrics according to their standard answer-matching or task-specific scoring rules. For captioning benchmarks, TextCaps and VATEX are evaluated with standard generation metrics, including CIDEr, BLEU-4, METEOR, and ROUGE-L, which measure caption quality from complementary perspectives such as n-gram precision, semantic overlap, and recall-oriented similarity. These metrics together assess answer correctness, caption generation quality, visual grounding accuracy, temporal reasoning, and long-context understanding.
\section{Additional Experiment Results}
\subsection{Additional Results on LongBench}
\label{appendix:llama}
Table~\ref{tab:longbench_llama} provides additional LongBench results with LLaMA under KV cache budgets of 1024 and 512. The results further support the generality of BACON beyond multimodal benchmarks. Across different compression backbones, BACON consistently improves the average performance over the corresponding base methods and MixKV-enhanced variants, showing that boundary-evidence calibration remains effective for long-context language understanding. The gains are particularly stable under the smaller budget of 512, where aggressive compression makes evidence retention more challenging. This suggests that BACON can better preserve tokens that become important near the generation boundary, rather than relying only on window-averaged attention signals.

We also observe that the improvements are not limited to a single task group. BACON achieves consistent gains on information localization tasks, such as QA over single-document and multi-document inputs, indicating that calibrated boundary evidence helps recover fine-grained relevant context. Meanwhile, it also improves many information aggregation and few-shot tasks, suggesting that the proposed calibration does not sacrifice global context modeling. A few metrics show smaller gains or occasional non-best results, especially in tasks where the baseline already performs close to the Full KV upper bound or where the score is saturated, such as passage retrieval. Overall, these appendix results demonstrate that BACON is not specific to visual-token compression, but provides a broadly applicable score calibration mechanism for KV cache compression.
% =========================================================
% Required packages
% =========================================================
% Put these in the preamble

\definecolor{promptrow}{RGB}{236,236,236}

% Centered X column
\newcolumntype{Y}{>{\centering\arraybackslash}X}

% =========================================================
% Direct Prompt Variations
% =========================================================
\begin{table*}[t]
\centering
\caption{\textbf{Robustness to direct prompt variations.}
Results under two alternative prompt templates (V1 and V2) on SnapKV, AdaKV, and SparseMM.
BACON consistently improves the corresponding base compressors across different prompt formulations.}
\label{tab:prompt_variations}
\vspace{-2mm}

\small
\setlength{\tabcolsep}{3.5pt}
\renewcommand{\arraystretch}{1.02}
\setlength{\aboverulesep}{0.15ex}
\setlength{\belowrulesep}{0.15ex}
\setlength{\cmidrulesep}{0.08ex}

\begin{tabularx}{0.97\textwidth}{
    >{\raggedright\arraybackslash}p{2cm}
    Y Y Y Y
    Y Y Y Y
}
\toprule
\multirow{2}{*}{\textbf{Method}}
& \multicolumn{4}{c}{\textbf{Prompt V1}}
& \multicolumn{4}{c}{\textbf{Prompt V2}} \\
\cmidrule(lr){2-5}
\cmidrule(lr){6-9}
& \textbf{DocVQA}
& \textbf{TextVQA}
& \textbf{ChartQA}
& \textbf{TextCaps}
& \textbf{DocVQA}
& \textbf{TextVQA}
& \textbf{ChartQA}
& \textbf{TextCaps} \\
\midrule

Full KV
& 93.50 & 82.60 & 70.60 & 1.48
& 93.70 & 83.20 & 71.40 & 0.44 \\

\midrule
\multicolumn{9}{l}{\textbf{Budget = 128}} \\

\textbf{SnapKV}
& 78.71 & 77.84 & 69.20 & 1.27
& 81.79 & 79.75 & 69.60 & 0.46 \\

\quad + MixKV
& 81.69 & 80.78 & 70.20 & 1.43
& 84.25 & \textbf{82.07} & \textbf{70.90} & 0.47 \\

\rowcolor{promptrow}
\quad + \textbf{BACON}
& \textbf{89.01} & \textbf{81.12} & \textbf{70.20} & \textbf{1.43}
& \textbf{91.14} & 82.04 & 70.60 & \textbf{0.61} \\

\midrule

\textbf{AdaKV}
& 79.56 & 76.28 & 69.30 & 1.19
& 82.34 & 78.11 & 69.50 & 0.44 \\

\quad + MixKV
& 81.80 & 78.99 & 69.80 & 1.37
& 84.27 & 81.13 & 70.60 & 0.43 \\

\rowcolor{promptrow}
\quad + \textbf{BACON}
& \textbf{89.08} & \textbf{80.64} & \textbf{70.10} & \textbf{1.44}
& \textbf{91.21} & \textbf{81.32} & \textbf{71.10} & \textbf{0.53} \\

\midrule

\textbf{SparseMM}
& 90.76 & 82.02 & 70.70 & 1.53
& 91.78 & 81.93 & \textbf{71.70} & 0.54 \\

\quad + MixKV
& 91.89 & 82.02 & 70.60 & 1.52
& \textbf{92.90} & 82.84 & 71.00 & 0.51 \\

\rowcolor{promptrow}
\quad + \textbf{BACON}
& \textbf{92.04} & \textbf{82.06} & \textbf{71.10} & \textbf{1.57}
& 92.88 & \textbf{83.21} & 71.60 & \textbf{0.58} \\

\midrule
\multicolumn{9}{l}{\textbf{Budget = 64}} \\

\textbf{SnapKV}
& 62.77 & 69.36 & 65.60 & 0.88
& 66.73 & 70.40 & 66.40 & 0.30 \\

\quad + MixKV
& 63.77 & 72.13 & 67.20 & 1.01
& 67.37 & 73.51 & 67.40 & 0.32 \\

\rowcolor{promptrow}
\quad + \textbf{BACON}
& \textbf{81.25} & \textbf{75.65} & \textbf{68.80} & \textbf{1.17}
& \textbf{84.14} & \textbf{78.87} & \textbf{69.90} & \textbf{0.35} \\

\midrule

\textbf{AdaKV}
& 63.29 & 67.52 & 65.60 & 0.86
& 67.39 & 69.69 & 66.80 & 0.27 \\

\quad + MixKV
& 64.27 & 69.69 & 67.30 & 0.94
& 68.04 & 72.05 & 67.50 & 0.31 \\

\rowcolor{promptrow}
\quad + \textbf{BACON}
& \textbf{81.95} & \textbf{75.03} & \textbf{69.00} & \textbf{1.18}
& \textbf{84.65} & \textbf{78.24} & \textbf{70.20} & \textbf{0.36} \\

\midrule

\textbf{SparseMM}
& 83.52 & 77.30 & 69.20 & 1.14
& 86.19 & 78.57 & 69.80 & 0.36 \\

\quad + MixKV
& 86.41 & 80.64 & 69.50 & 1.42
& 88.26 & \textbf{81.54} & 70.50 & 0.41 \\

\rowcolor{promptrow}
\quad + \textbf{BACON}
& \textbf{90.98} & \textbf{81.08} & \textbf{70.10} & \textbf{1.47}
& \textbf{92.12} & 81.52 & \textbf{70.50} & \textbf{0.49} \\

\bottomrule
\end{tabularx}

\vspace{-3mm}
\end{table*}

% =========================================================
% Two-Stage CoT
% =========================================================
\begin{table*}[t]
\centering
\caption{\textbf{Robustness under two-stage chain-of-thought prompting.}
Results on SnapKV, AdaKV, and SparseMM under KV-cache budgets of 64 and 128.
BACON remains effective under the substantially different two-stage CoT inference procedure.}
\label{tab:prompt_cot}
\vspace{-2mm}

\small
\setlength{\tabcolsep}{4.8pt}
\renewcommand{\arraystretch}{1.02}
\setlength{\aboverulesep}{0.15ex}
\setlength{\belowrulesep}{0.15ex}
\setlength{\cmidrulesep}{0.08ex}

\begin{tabular}{l cccc cccc}
\toprule
\multirow{2}{*}{\textbf{Method}}
& \multicolumn{4}{c}{\textbf{Budget 64}}
& \multicolumn{4}{c}{\textbf{Budget 128}} \\
\cmidrule(lr){2-5}
\cmidrule(lr){6-9}
& \textbf{DocVQA}
& \textbf{TextVQA}
& \textbf{ChartQA}
& \textbf{TextCaps}
& \textbf{DocVQA}
& \textbf{TextVQA}
& \textbf{ChartQA}
& \textbf{TextCaps} \\
\midrule

Full KV
& 93.30 & 77.60 & 68.90 & 0.67
& 93.30 & 77.60 & 68.90 & 0.67 \\

\midrule

\textbf{SnapKV}
& 55.79 & 55.13 & 54.80 & 0.11
& 84.41 & 74.62 & 64.90 & 0.66 \\

\quad + MixKV
& 56.26 & 55.42 & 56.20 & 0.10
& 84.08 & 75.97 & 64.70 & 0.76 \\

\rowcolor{promptrow}
\quad + \textbf{BACON}
& \textbf{62.06} & \textbf{56.75} & \textbf{56.20} & \textbf{0.61}
& \textbf{84.79} & \textbf{76.44} & \textbf{65.60} & \textbf{0.84} \\

\midrule

\textbf{AdaKV}
& 57.79 & 53.30 & 54.80 & 0.12
& 81.93 & 71.01 & 64.40 & 0.67 \\

\quad + MixKV
& 58.27 & \textbf{54.29} & \textbf{59.40} & 0.12
& 82.50 & 73.34 & \textbf{64.70} & 0.74 \\

\rowcolor{promptrow}
\quad + \textbf{BACON}
& \textbf{59.90} & 54.10 & 58.30 & \textbf{0.57}
& \textbf{82.62} & \textbf{73.88} & 64.50 & \textbf{0.82} \\

\midrule

\textbf{SparseMM}
& 83.67 & 72.99 & 60.70 & 0.32
& 91.37 & 77.46 & 66.10 & 0.79 \\

\quad + MixKV
& 83.99 & 73.92 & 59.90 & 0.25
& 90.39 & 76.67 & 66.10 & 0.78 \\

\rowcolor{promptrow}
\quad + \textbf{BACON}
& \textbf{84.65} & \textbf{74.19} & \textbf{63.00} & \textbf{0.64}
& \textbf{91.70} & \textbf{78.78} & \textbf{66.10} & \textbf{0.92} \\

\bottomrule
\end{tabular}

\vspace{-3mm}
\end{table*}
\subsection{Results on Extreme Multimodal Context}
\label{append:extreme} 
We further evaluate BACON under two extreme multimodal scenarios with substantially enlarged visual contexts: ultra-high-resolution image understanding on HR-Bench and long-form video understanding on Video-MME. We use Qwen2-VL-7B-Instruct following the same evaluation protocol as the main experiments. For HR-Bench, we evaluate both 4K and 8K settings with a KV-cache budget of 128. For Video-MME, we use a KV-cache budget of 1024 and restrict evaluation to 30--60 minute videos without subtitles. \paragraph{Ultra-high-resolution images.} As shown in Table~\ref{tab:extreme_scenarios}, BACON remains effective when the visual context grows substantially due to high-resolution image tiling. On HR-Bench 4K, BACON improves SnapKV and SparseMM from 63.9 to 64.3, matching the Full KV result. On the more challenging 8K setting, BACON improves SnapKV from 61.5 to 61.8 and improves both AdaKV and SparseMM from 61.5 to 62.1, again reaching Full KV performance for the latter two compressors. These results indicate that BACON can preserve sparse answer-relevant evidence even when it is distributed over substantially longer high-resolution visual sequences. \paragraph{Long-form videos.} BACON also remains effective under extreme temporal contexts. On 30--60 minute Video-MME videos, it improves SnapKV and AdaKV from 51.7 to 52.3, and improves SparseMM from 51.7 to 53.0. In particular, SparseMM + BACON approaches the Full KV result of 53.3 despite operating with a compressed KV cache. This suggests that boundary evidence remains informative for identifying sparse task-relevant content even when the visual context spans long temporal sequences. Overall, BACON maintains its effectiveness as visual context length increases along both spatial and temporal dimensions. The results on high-resolution images and long videos further demonstrate that its evidence-preservation mechanism generalizes beyond the standard multimodal settings considered in the main experiments.
\subsection{Additional Ablation Results}
\label{append:ablation}
We provide additional ablation results in Tables~\ref{tab:appendix_ablation_llava_b128} and~\ref{tab:appendix_ablation_qwen} to further validate the component design of BACON beyond the main setting. The full BACON variant consistently achieves the best or near-best performance across different models, compression backbones, benchmarks, and cache budgets, confirming the complementary roles of boundary evidence $E$, local coherence $L$, and cross-layer trace support $T$. Removing any component often weakens performance or leads to less stable results, especially on evidence-sensitive tasks such as DocVQA, TextVQA, and TextCaps. These additional results further demonstrate that BACON's component design remains robust across broader model and budget settings.
\subsection{Prompt Robustness Analysis}
\label{append:prompt}

We further evaluate whether BACON is sensitive to the prompt formulation or inference procedure. In addition to the default prompt used in the main experiments, we consider two semantically equivalent direct prompt variants and a two-stage chain-of-thought (CoT) setting. Prompt Variation 1 places the task instruction before the question, whereas Prompt Variation 2 presents the question before the output constraint. Both variants preserve the original task semantics and required answer format, while altering the ordering and surface realization of the prompt. For two-stage CoT, the model first generates image-grounded intermediate reasoning and subsequently produces the final answer conditioned on this reasoning in a separate generation stage; only the final answer is used for evaluation.

Table~\ref{tab:prompt_variations} reports the results under the two direct prompt variations. Across different compressors and KV-cache budgets, BACON generally improves the corresponding base compressor and remains competitive with or superior to MixKV across most settings. The improvements are particularly pronounced under the more aggressive budget of 64, where compression is more likely to discard sparse answer-relevant visual evidence. These results indicate that the effectiveness of BACON does not depend on a particular ordering or wording of the input prompt.

We further evaluate BACON under the substantially different two-stage CoT procedure in Table~\ref{tab:prompt_cot}. Despite the additional intermediate reasoning stage and the resulting change in the inference process, BACON continues to provide overall improvements across compressors and datasets, with the strongest benefits again appearing under tighter KV budgets. This suggests that BACON can preserve visual evidence that remains useful not only for direct answer generation, but also when the model must first construct intermediate reasoning before producing its final prediction.

Overall, the results across direct prompt reformulations and two-stage reasoning show that BACON is not tied to a specific prompt template or decoding procedure. Its gains remain robust under changes in prompt structure and inference behavior, supporting the generality of its evidence-preservation mechanism under aggressive KV cache compression.

\subsection{Additional Visualization Results}
\label{append:vis}
Here we provide more visualization results on how BACON improve visual evidence retention compared to baseline methods, as shown in Fig.~\ref{appendix:vis}
\begin{figure*}[t]
    \centering
    \includegraphics[width=0.95\textwidth]{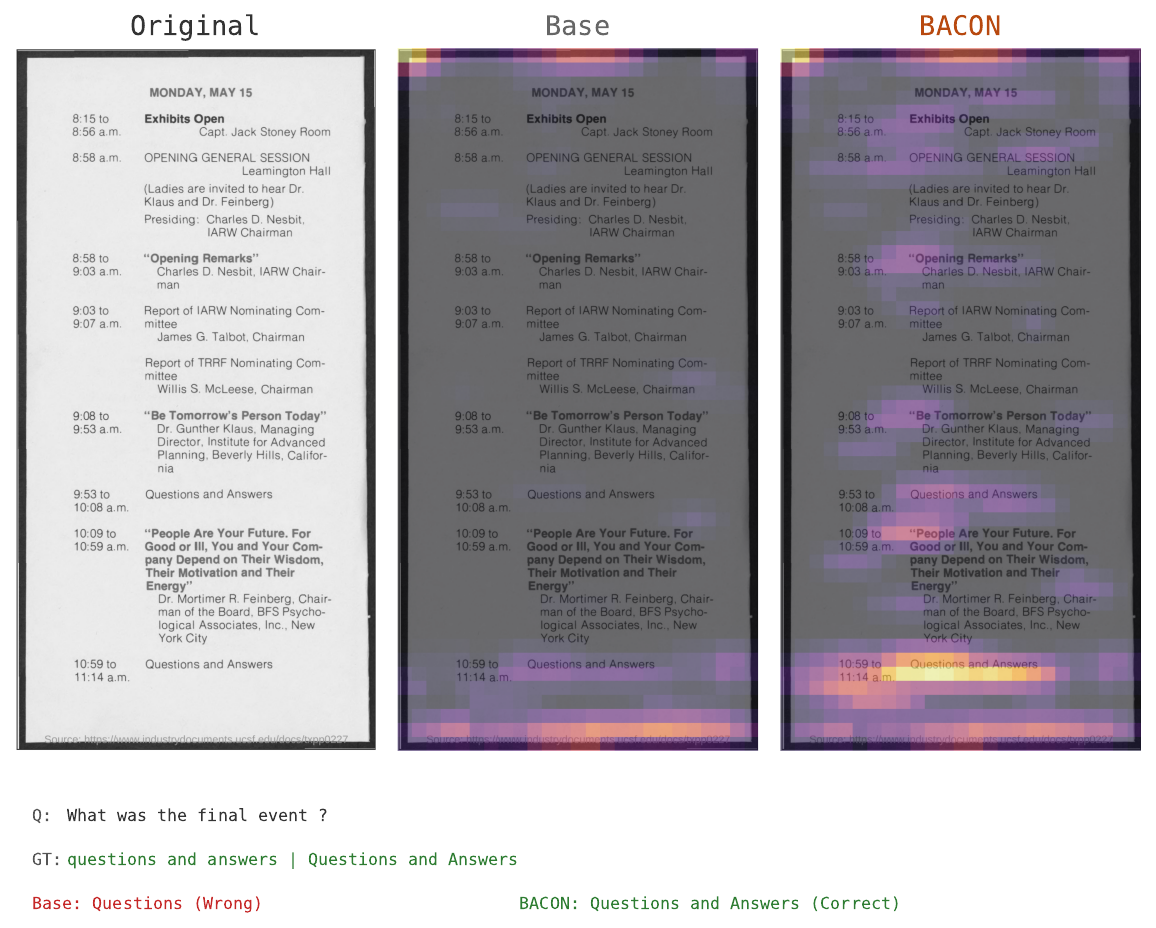}
    \caption{Additional visualizations comparing evidence importance estimation between window attention and BACON.}
    \label{appendix:vis}
\end{figure*}

\begin{figure*}[t]
    \ContinuedFloat
    \centering
    \includegraphics[width=0.95\textwidth]{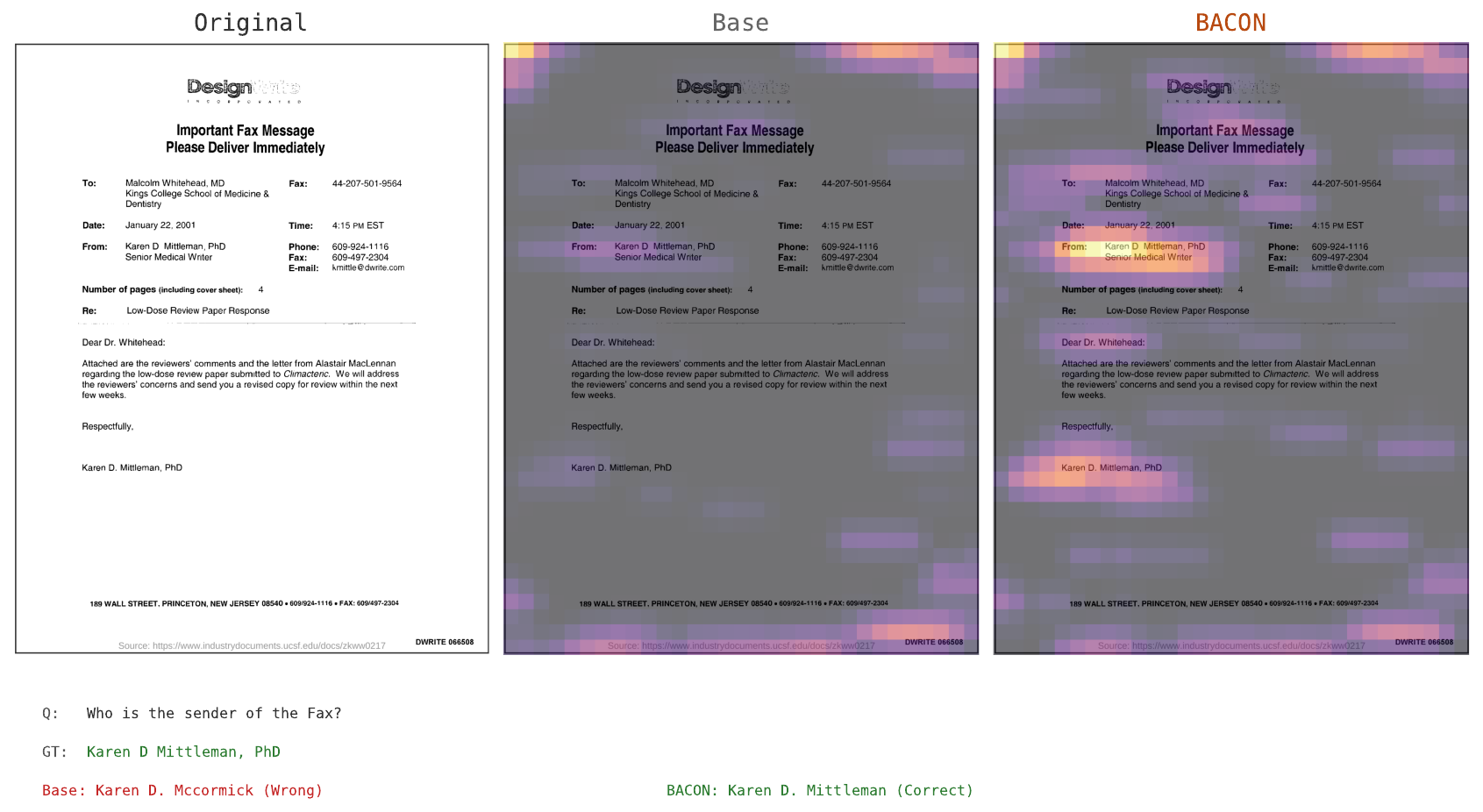}
    \caption{Additional visualizations comparing evidence importance estimation between window attention and BACON.}
\end{figure*}

\begin{figure*}[t]
    \ContinuedFloat
    \centering
    \includegraphics[width=0.95\textwidth]{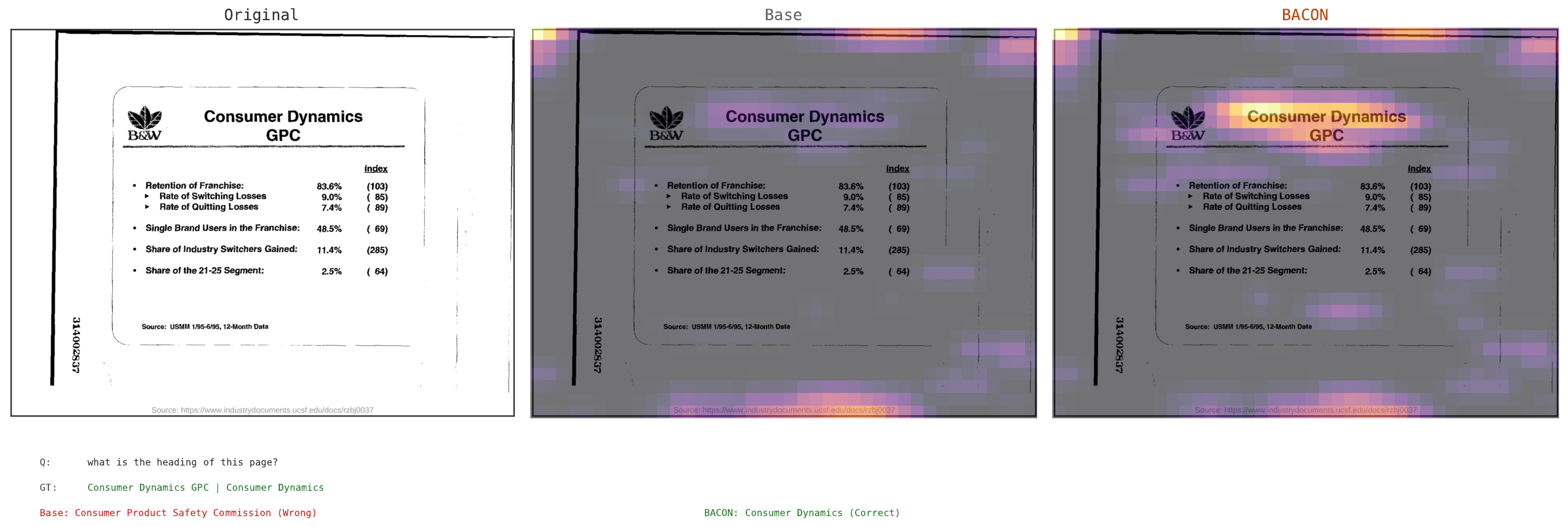}
    \caption{Additional visualizations comparing evidence importance estimation between window attention and BACON.}
\end{figure*}

\begin{figure*}[t]
    \ContinuedFloat
    \centering
    \includegraphics[width=0.95\textwidth]{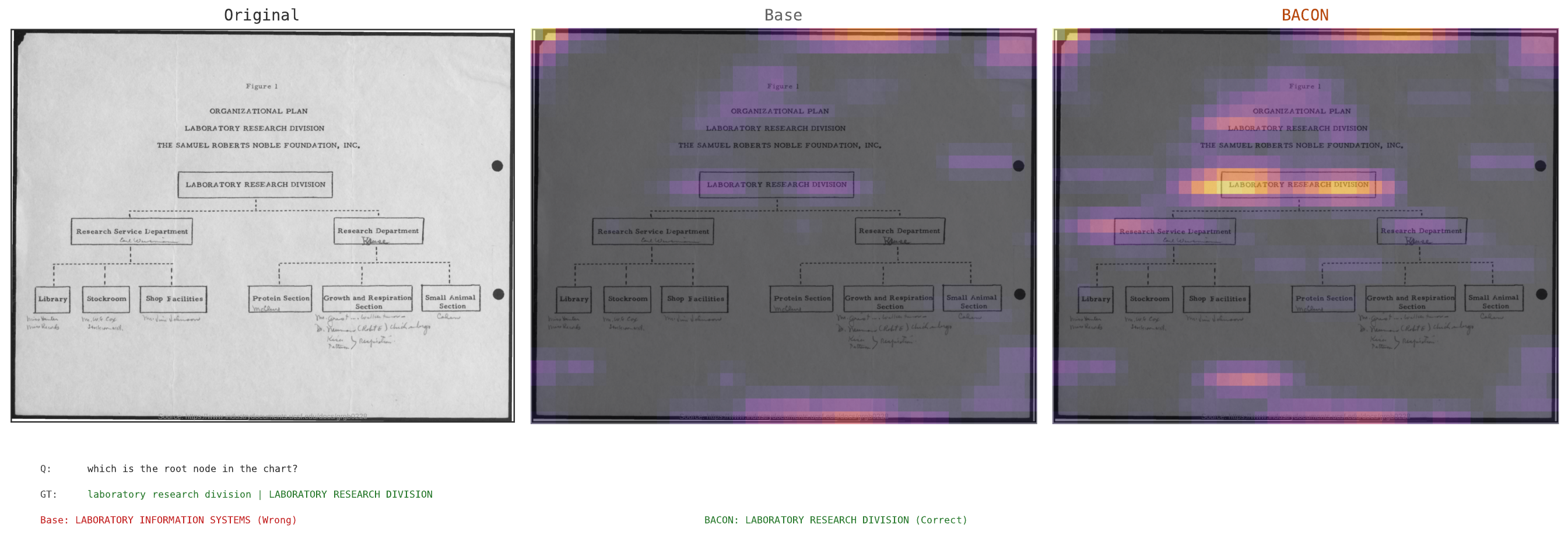}
    \caption{Additional visualizations comparing evidence importance estimation between window attention and BACON.}
\end{figure*}

\end{document}